\documentclass[sigconf]{aamas}

\usepackage{balance} 
\usepackage{bm}
\usepackage{subfigure}
\usepackage{multirow}
\usepackage{algorithm}
\usepackage{algpseudocode}  
\usepackage{balance}

\doi{JWPH6906}

\makeatletter
\gdef\@copyrightpermission{
  \begin{minipage}{0.2\columnwidth}
   \href{https://creativecommons.org/licenses/by/4.0/}{\includegraphics[width=0.90\textwidth]{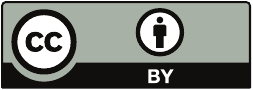}}
  \end{minipage}\hfill
  \begin{minipage}{0.8\columnwidth}
   \href{https://creativecommons.org/licenses/by/4.0/}{This work is licensed under a Creative Commons Attribution International 4.0 License.}
  \end{minipage}
  \vspace{5pt}
}
\makeatother

\setcopyright{ifaamas}
\acmConference[AAMAS '26]{Proc.\@ of the 25th International Conference
on Autonomous Agents and Multiagent Systems (AAMAS 2026)}{May 25 -- 29, 2026}
{Paphos, Cyprus}{C.~Amato, L.~Dennis, V.~Mascardi, J.~Thangarajah (eds.)}
\copyrightyear{2026}
\acmYear{2026}
\acmDOI{}
\acmPrice{}
\acmISBN{}

\title[AAMAS-2026 Formatting Instructions]{Neuro-symbolic Action Masking for Deep Reinforcement Learning}

\author{Shuai Han}
\affiliation{
	\institution{Utrecht University}
	\city{Utrecht}
	\country{the Netherland}}
\email{s.han@uu.nl}

\author{Mehdi Dastani}
\affiliation{
	\institution{Utrecht University}
	\city{Utrecht}
	\country{the Netherland}}
\email{m.m.dastani@uu.nl}

\author{Shihan Wang}
\affiliation{
	\institution{Utrecht University}
	\city{Utrecht}
	\country{the Netherland}}
\email{s.wang2@uu.nl}

\begin{abstract}
Deep reinforcement learning (DRL) may explore infeasible actions during training and execution. Existing approaches assume a symbol grounding function that maps high-dimensional states to consistent symbolic representations and a manually specified action masking techniques to constrain actions. In this paper, we propose Neuro-symbolic Action Masking (NSAM), a novel framework that automatically learn symbolic models, which are consistent with given domain constraints of high-dimensional states, in a minimally supervised manner during the DRL process. Based on the learned symbolic model of states, NSAM learns action masks that rules out infeasible actions. NSAM enables end-to-end integration of symbolic reasoning and deep policy optimization, where improvements in symbolic grounding and policy learning mutually reinforce each other. We evaluate NSAM on multiple domains with constraints, and experimental results demonstrate that NSAM significantly improves sample efficiency of DRL agent while substantially reducing constraint violations.
\end{abstract}


\newcommand{\BibTeX}{\rm B\kern-.05em{\sc i\kern-.025em b}\kern-.08em\TeX}

\begin{document}


\pagestyle{fancy}
\fancyhead{}


\maketitle 


\section{Introduction}

With the powerful representation capability of neural networks, deep reinforcement learning (DRL) has achieved remarkable success in a variety of complex domains that require autonomous agents, such as autonomous driving \cite{autodriving4_1, autodriving4_2, autodriving4_3}, resource management \cite{resourcem4_1, resourcem4_2}, algorithmic trading \cite{autotrading4_1, autotrading4_2} and robotics \cite{roboticRL4_1, roboticRL4_2, roboticRL4_3}. However, in real-world scenarios, agents face the challenges of learning policies from few interactions \cite{roboticRL4_2} and keeping violations of domain constraints to a minimum during training and execution \cite{autodriving_safe}. To address these challenges, an increasing number of neuro-symbolic reinforcement learning (NSRL) approaches have been proposed, aiming to exploit the structural knowledge of the problem to improve sample efficiency \cite{shindo2024blendrl,RM,nsrl2025_planning} or to constrain agents to select actions \cite{PLPG,PPAM,nsrl2024_plpg_multi}. 

Among these NSRL approaches, a promising practice is to exclude infeasible actions for the agents.\footnote{We use the term infeasible actions throughout the paper, which can also be considered as unsafe, unethical or in general undesirable actions.}  This is typically achieved by assuming a predefined symbolic grounding \cite{nsplanning} or label function \cite{RM} that maps high-dimensional states into symbolic representations and manually specify action masking techniques \cite{actionmasking_app1, actionmasking_app3, actionmasking_app4}. However, predefining the symbolic grounding function is often expensive \cite{neuroRM}, as it requires complete knowledge of the environmental states, and could be practically impossible when the states are high-dimensional or infinite. Learning symbolic grounding from environmental state is therefore crucial for NSRL approaches and remains a highly challenging problem \cite{neuroRM}.

In particular, there are three main challenges. First, real-world environments should often satisfy complex constraints expressed in a domain specific language, which makes learning the symbolic grounding function difficult \cite{ahmed2022semantic}. Second, obtaining full supervision for learning symbolic representations in DRL environments is unrealistic, as those environments rarely provide the ground-truth symbolic description of every state. Finally, even if symbolic grounding can be learned, integrating it into reinforcement learning to achieve end-to-end learning remains a challenge.

To address these challenges, we propose Neuro-symbolic Action Masking (NSAM), a framework that integrates symbolic reasoning into deep reinforcement learning. The basic idea is to use probabilistic sentential decision diagrams (PSDDs) to learn symbolic grounding. PSDDs serve two purposes: they guarantee that any learned symbolic model satisfies domain constraints expressed in a domain specific language \cite{kisa2014probabilistic}, and they allow the agent to represent probability distributions over symbolic models conditioned on high-dimensional states. In this way, PSDDs bridge the gap between numerical states and symbolic reasoning without requiring manually defined mappings. Based on the learned PSDDs, NSAM combines action preconditions with the inferred symbolic model of numeric states to construct action masks, thereby filtering out infeasible actions. Crucially, this process only relies on minimal supervision in the form of action explorablility feedback, rather than full symbolic description at every state. Finally, NSAM is trained end-to-end, where the improvement of symbolic grounding and policy optimization mutually reinforce each other.

We evaluate NSAM on four DRL decision-making domains with domain constraints, and compare it against a series of state-of-the-art baselines. Experimental results demonstrate that NSAM not only learns more efficiently, consistently surpassing all baselines, but also substantially reduces constraint violations during training. The results further show that the symbolic grounding plays a crucial role in exploiting underlying knowledge structures for DRL.

\section{Problem setting}
\label{NSAM:probsetting}

We study reinforcement learning (RL) on a Markov
Decision Process (MDP) \cite{RL1998} $\mathcal{M} = (\mathcal{S}, \mathcal{A}, \mathcal{T}, R, \gamma)$ where $\mathcal{S}$ is a set of states, $\mathcal{A}$ is a finite set of actions, $\mathcal{T}: \mathcal{S} \times \mathcal{A} \times \mathcal{S} \rightarrow [0, 1]$ is a transition function, $\gamma \in [0, 1)$ is a discount factor and $R: \mathcal{S} \times \mathcal{A} \times \mathcal{S} \rightarrow \mathbb{R}$ is a reward function. An agent employs a policy $\pi$ to interact with the environment. At a time step $t$, the agent takes action $a_t$ according to the current state $s_t$. The environment state will transfer to next state $s_{t+1}$ based on the transition probability $\mathcal{T}$. The agent will receive the reward $r_t$. Then, the next round of interaction begins. The goal of this agent is to find the optimal policy $\pi^*$ that maximizes the expected return: $\mathbb{E}[\sum_{t=0}^{T}\gamma^t r_t | \pi]$, where $T$ is the terminal time step.

To augment RL with symbolic domain knowledge, we extend the normal MDP with the following modules $(\mathcal{P}, \mathcal{AP}, \phi)$ where $\mathcal{P} = \{p_1,..,p_K\}$ is a finite set of atomic propositions (each $p \in \mathcal{P}$ represents a Boolean property of a state $s \in \mathcal{S}$ ), $\mathcal{AP} = \{(a, \varphi)| a\in \mathcal{A}, \varphi \in L(\mathcal{P}) \}$ is the set of actions with their preconditions, and $L(\mathcal{P})$ denotes the propositional language over $\mathcal{P}$. We use $(a, \varphi)$ to state that action $a$ is \textit{explorable}\footnote{All actions $a \in \mathcal{A}$ can in principle be chosen by the agent. However, we use the term \textit{explorable} to distinguish actions whose preconditions are satisfied (safe, ethical, desriable actions) from those whose preconditions are not satisfied (unsafe, unethical, undesirable actions).} in a state if and only if its precondition $\varphi$ holds in that state, $\phi \in L(\mathcal{P})$ is a domain constraint. We use  $|[\phi]| = \{\bm{m} | \bm{m} \models \phi\}$ to denote the set of all possible symbolic models of $\phi$~\footnote{a model is a truth assignment to all propositions in $\mathcal{P}$}.

\begin{figure}[t]
	\centering
	\subfigure[ ]{\includegraphics[width=2.1cm]{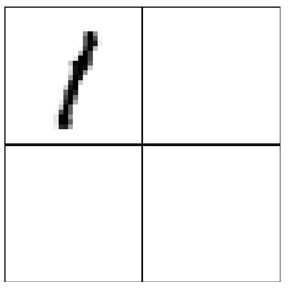}\label{fig:sudo1}}
	\subfigure[ ]{\includegraphics[width=2.1cm]{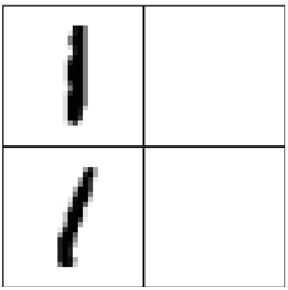}\label{fig:sudo2}}
	\subfigure[ ]{\includegraphics[width=2.1cm]{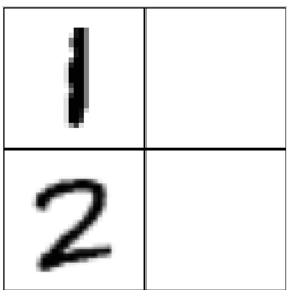}\label{fig:sudo3}}
	\vspace{-0.4cm}
	\caption{Example states in the Visual Sudoku environment}
	\label{fig:example_sudo}
\end{figure}

To illustrate how symbolic domain knowledge $(\mathcal{P}, \mathcal{AP}, \phi)$ is reflected in our formulation, we consider the Visual Sudoku task as a concrete example. In this environment, each state is represented as a non-symbolic image input. The properties of a state can be described using propositions in $\mathcal{P}$. For example, the properties of the state in Figure \ref{fig:sudo1} include `position (1,1) is number 1', `position (1,2) is empty', etc. Each action $a$ of filling a number in a certain position corresponds to a symbolic precondition $\varphi$, represented by $(a, \varphi) \in \mathcal{AP}$. For example, the action `filling number 1 at position (1,1)' requires that both propositions `position (1,2) is number 1' and `position (2,1) is number 1' are false. Finally, $\phi$ is used to constrain the set of possible states, e.g., `position (1,1) is number 1' and `position (1,1) is number 2' cannot both be simultaneously true for a given state. To leverage this knowledge, challenges arise due to the following problems.

\noindent\textbf{(P1) Numerical–symbolic gap.} Knowledge is based on symbolic property of states, but only raw numerical states are available.


\noindent\textbf{(P2) Constraint satisfaction.} The truth values of propositions in $\mathcal{P}$ mapped from a DRL state $s$ must satisfy domain constraints $\phi$. 

\begin{figure*}[t]
	\centering
	\subfigure[Distribution]{\includegraphics[width=2.2cm]{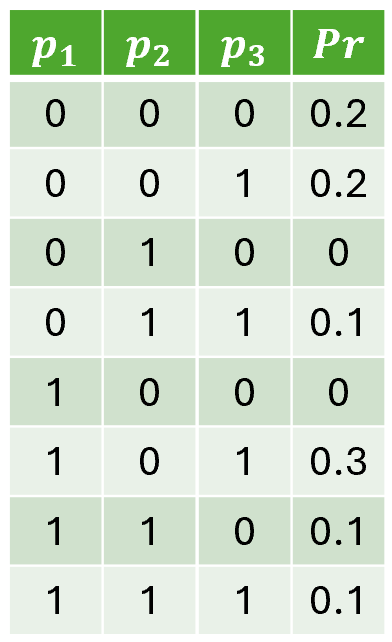}\label{fig:psdd_dis}}
	\subfigure[SDD]{\includegraphics[width=3.3cm]{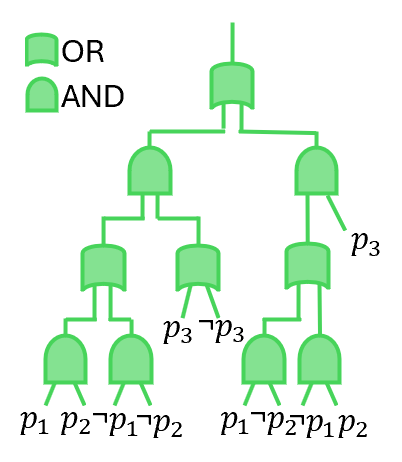}\label{fig:psdd_sdd}}
	\subfigure[PSDD]{\includegraphics[width=3.2cm]{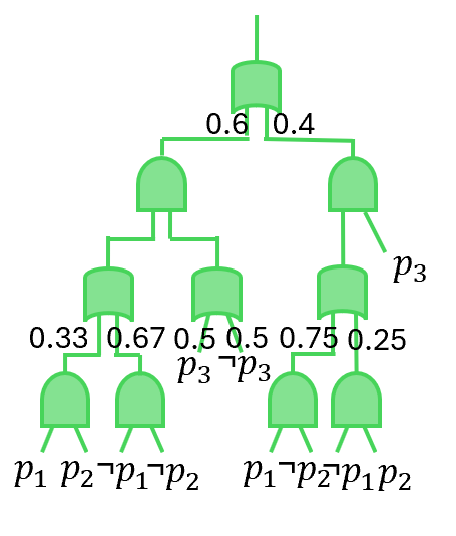}\label{fig:psdd_psdd}}
	\subfigure[Vtree]{\includegraphics[width=3.6cm]{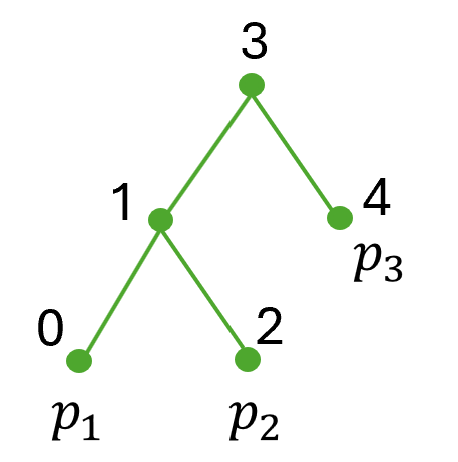}\label{fig:psdd_vtree}}
	\subfigure[A general fragment]{\includegraphics[width=3.6cm]{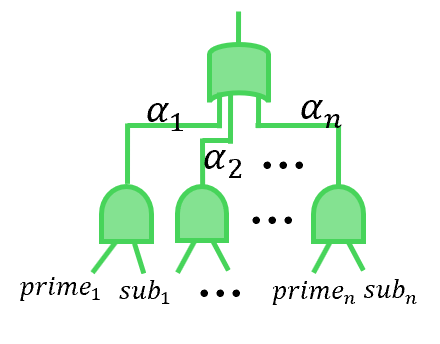}\label{fig:psdd_fragment}}
	\vspace{-0.3cm}
	\caption{(a) An example of joint distribution for three propositions $p_1, p_2$ and $p_3$ with the constraint $(p_1 \leftrightarrow p_2) \lor p_3$. (b) A SDD circuit with `OR' and `AND' logic gate to represent the constrain $(p_1 \leftrightarrow p_2) \lor p_3$. (c) The PSDD circuit to represent the distribution in Fig. \ref{fig:psdd_dis}. (d) The vtree used to group variables. (e) A general fragment to show the structure of SDD and PSDD.} 
	\label{fig:psdd}
\end{figure*}

\noindent\textbf{(P3) Minimal supervision.} The RL environment cannot provide full ground truth of propositions at each state.

\noindent\textbf{(P4) Differentiability.} The symbolic reasoning with $\varphi$ introduces non-differentiable process, which could be conflicting with gradient-based DRL algorithms that require differentiable policies.

\noindent\textbf{(P5) End-to-end learning.} Achieving end-to-end training on prediction of propositions, symbolic reasoning over preconditions and optimization of policy is challenging.

In summary, the above challenges fall into three categories. (P1–P3) concern learning symbolic models from high-dimensional states in DRL, which we address in Section \ref{NSAM:sec3}. (P4) relates to the differentiability barrier when combining symbolic reasoning with gradient-based DRL, which we tackle in Section \ref{NSAM:sec4}. (P5) raises the need for an end-to-end training, which we present in Section \ref{NSAM:sec_meth05}.

\section{Learning Symbolic Grounding}
\label{NSAM:sec3}

This section introduces how NSAM learns symbolic grounding. At a high level, the goal is to learn whether an action is explorable in a state. Specifically, the agent receives minimal supervision from state transitions after executing an action $a$. Using this supervision, NSAM learns to estimate the symbolic model of the high-dimensional input state that is in turn used to check the satisfiability of action preconditions. To achieve this, Section~\ref{NSAM_section:psdd} presents a knowledge compilation step to encode domain constraints into a symbolic structure, while Section~\ref{NSAM_section:learnPSDD} explains how this symbolic structure is parameterized and learned from minimal supervision.

\subsection{Compiling the Knowledge}
\label{NSAM_section:psdd}

To address P2 (Constraint satisfaction), we introduce the Probabilistic Sentential Decision Diagram (PSDD) \cite{kisa2014probabilistic}. PSDDs are designed to represent probability distributions $Pr(\bm{m})$ over possible models, where any model $\bm{m}$ that violates domain constraints is assigned zero probability \cite{conditionalPSDD}. For example, consider the distribution in Figure 2(a). The first step in constructing a PSDD is to build a Boolean circuit that captures the entries whose probability values are always zero, as shown in Figure 2(b). Specifically, the circuit evaluates to $0$ for model $\bm{m}$ if and only if $\bm{m}\not\models \phi$. The second step is to parameterize this Boolean circuit to represent the (non-zero) probability of valid entries, yielding the PSDD in Figure 2(c).


To obtain the Boolean circuit in Figure 2(b), we represent the domain constraint $\phi$ using a general data structure called a Sentential Decision Diagram (SDD) \cite{sdd}. An SDD is a normal form of a Boolean formula that generalizes the well-known Ordered Binary Decision Diagram (OBDD) \cite{OBDD,OBDD2}. SDD circuits satisfy specific syntactic and semantic properties defined with respect to a binary tree, called a vtree, whose leaves correspond to propositions (see Figure 2(d)). Following Darwiche’s definition \cite{sdd, psdd_infer1}, an SDD normalized for a vtree $v$ is a Boolean circuit defined as follows: If $v$ is a leaf node labeled with variable $p$, the SDD is either $p$, $\neg p$, $\top$, $\bot$, or an OR gate with inputs $p$ and $\neg p$. If $v$ is an internal node, the SDD has the structure shown in Figure 2(e), where $\textit{prime}_1,\ldots,\textit{prime}_n$ are SDDs normalized for the left child $v^l$, and $\textit{sub}_1,\ldots,\textit{sub}_n$ are SDDs normalized for the right child $v^r$. SDD circuits alternate between OR gates and AND gates, with each AND gate having exactly two inputs. The OR gates are mutually exclusive in that at most one of their inputs evaluates to true under any circuit input \cite{sdd, psdd_infer1}.

A PSDD is obtained by annotating each OR gate in an SDD with parameters $(\alpha_1, \ldots, \alpha_n)$ over its inputs \cite{kisa2014probabilistic, psdd_infer1}, where $\sum_i \alpha_i = 1$ (see Figure 2(e)). The probability distribution defined by a PSDD is as follows. Let $\bm{m}$ be a model that assigns truth values to the PSDD variables, and suppose the underlying SDD evaluates to $0$ under $\bm{m}$; then $Pr(\bm{m}) = 0$. Otherwise, $Pr(\bm{m})$ is obtained by multiplying the parameters along the path from the output gate.

The key advantage of using PSDDs in our setting is twofold. First, PSDDs strictly enforce domain constraints by assigning zero probability to any model $\bm{m}$ that violates $\phi$ \cite{conditionalPSDD}, thereby ensuring logical consistency (P2). Second, by ruling out impossible truth assignment through domain knowledge, PSDDs effectively reduce the scale of the probability distribution to be learned \cite{ahmed2022semantic}. 

Besides, PSDDs also support tractable probabilistic queries \cite{PCbooks, psdd_infer1}. While PSDD compilation can be computationally expensive as its size grows exponentially in the number of propositions and constraints, it is a one-time offline cost. Once compilation is completed, PSDD inference is linear-time, making symbolic reasoning efficient during both training and execution \cite{psdd_infer1}. 

\subsection{Learning the parameters of PSDD in DRL}
\label{NSAM_section:learnPSDD}
\begin{figure}[t]
	\centering
	\includegraphics[width=8cm]{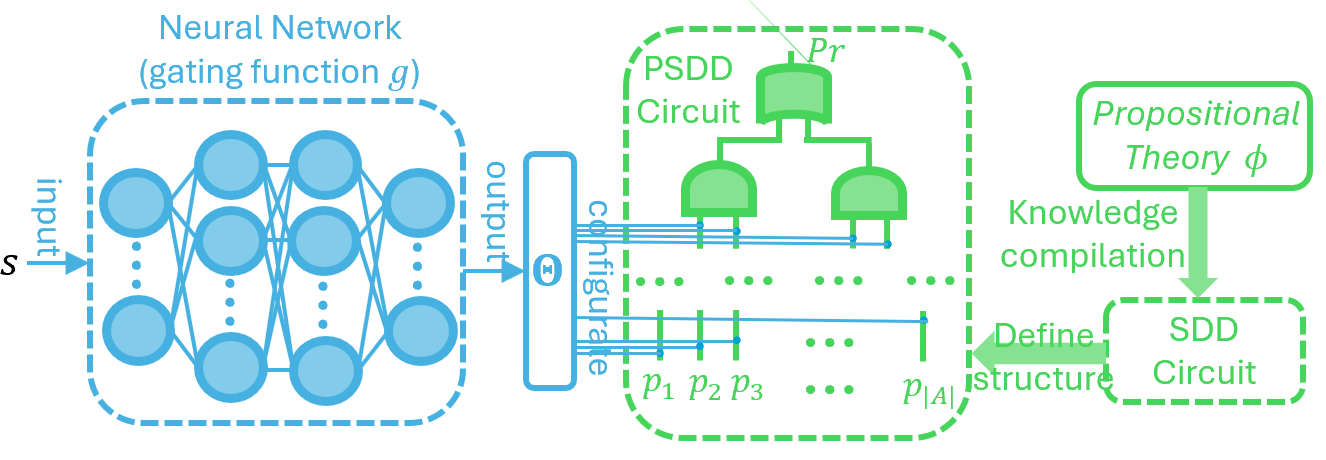}
	\vspace{-0.2cm}
	\caption{The architecture design to calculate the probability of symbolic model $\bm{m}$ given DRL state $s$.}
	\label{fig:Framework}
\end{figure}

To address P1 (Numerical–symbolic gap), we need to learn distributions of models that satisfy the domain constraints. Inspired by recent deep supervised learning work on PSDDs \cite{ahmed2022semantic}, we parameterize the PSDD using the output of gating function $g$. This gating function is a neural network that maps high-dimensional RL states to PSDD parameters $\Theta = g(s)$. This design allows the PSDD to represent state-conditioned distributions over propositions through its learned parameters, while strictly adhering to domain constraints (via its structure defined by symbolic knowledge $\phi$). The overall process is shown in Figure~\ref{fig:Framework}. We use $Pr(\bm{m} \mid \bm{\Theta} = g(s), \bm{m}\models\phi)$ to denote the probability of model $\bm{m}$ that satisfy the domain constrains $\phi$ given the state $s$ (this is calculated by PSDD in Figure~\ref{fig:Framework}).   


After initializing $g$ and the PSDD according to the structure in Figure~\ref{fig:Framework}, we obtain a distribution over $\bm{m}$ such that for all $\bm{m} \not\models \phi$, $Pr(\bm{m} \mid \bm{\Theta} = g(s), \bm{m}\not\models\phi)=0$. However, for the probability distribution over $\bm{m}$ that does satisfy $\phi$, we still need to learn from data to capture the probability of different $\bm{m}$ by adjusting parameters of gating function $g$. To train the PSDD from minimal supervision signals (for problem (P3) in Section \ref{NSAM:probsetting}), we construct the supervision data from $\Gamma_{\phi}$, which consists of tuples $(s,a,s',y)$ where transitions $(s,a,s')$ are explored from the environment and $y$ is calculated by:
\begin{equation} \label{NSAM:DATA}
	y = 
	\begin{cases}
		1, & \text{if } s \;\text{and}\;s'\;\text{do not violate}\; \phi,\\
		0, & \text{otherwise.}
	\end{cases}
\end{equation}
That is, the action $a$ is labeled as explorable (i.e., $y=1$) in state $s$ if it does not lead to a violation of the domain constraint $\phi$; otherwise the action a is not explorable (i.e., $y=0$).

Unlike a fully supervised setting that expensively requires labeling every propositional variable in $P$, Eq.~(\ref{NSAM:DATA}) only requires labeling whether a given state violates the domain constraint $\phi$, which is a minimal supervision signal. In practice, the annotation of $y$ can be obtained either (i) by providing labeled data on whether the resulting state $s'$ violates the constraint $\phi$~\cite{book2006}, or (ii) via an automatic constraint-violation detection mechanism~\cite{autochecking1,autochecking2}.

We emphasize that action preconditions $\varphi$ and the domain constraints $\phi$ are two separate elements and treated differently. We first automatically generate training data to learn PSDD parameters by constructing tuples $(s,a,s',y)$ as defined in Equation~(\ref{NSAM:DATA}). The argument $y$ in tuples $(s,a,s’,y)$ is then used as an indicator for action preconditions. Specifically, we use $y$ to label whether action $a$ is excutable in state $s$, i.e., if transition $(s,a,s')$ is explored by DRL policy in a non-violating states $s$ and $s'$, then $y=1$, meaning that action a is explorable in $s$; otherwise $y=0$. We thus use $y$ in $(s,a,s’,y)$ as a minimal supervision signal to estimate the probability of the precondition of action $a$ being satisfied in non-violating $s$ during PSDD training.

By continuously rolling out the DRL agent in the environment, we store $(s,a,s',y)$ into a buffer $\mathcal{D}$. After collecting sufficient data, we sample batches from $\mathcal{D}$ and update $g$ via stochastic gradient descent~\cite{SDG,ADAM}. Concretely, the update proceeds as follows. Given samples $(s, a, s',y)$, we first use the current PSDD to estimate the probability that action $a$ is explorable in state $s$, i.e., the probability that $s$ satisfies the precondition $\varphi$ associated with $a$ in $\mathcal{AP}$:
\begin{equation}\label{NSAM:hatp}
	\hat{P}(a|s) = \sum_{\bm{m}\models \varphi} Pr(\bm{m}|\bm{\Theta}=g(s), \bm{m}\models\phi)
\end{equation}
Note that $\hat{P}(a|s)$ here does not represent a policy as in standard DRL; rather, it denotes the estimated probability that action $a$ is explorable in state $s$. As shown in Equation (\ref{NSAM:hatp}), this probability is calculated by aggregating the probabilities of all models $\bm{m}$ that satisfy the precondition $\varphi$. In addition, to evaluate if $\bm{m} \models \varphi$, we assign truth values to the leaf variables of $\varphi$'s SDD circuit based on $\bm{m}$ and propagate them bottom-up through the `OR' and `AND'  gates, where the Boolean value at the root indicates the satisfiability.

Given the probability estimated from Equation (\ref{NSAM:hatp}), we compute the cross-entropy loss \cite{CROSSENTR} by comparing it with the explorability label $y$. Specifically, for a single data $(s,a,s',y)$, the loss is:
\begin{equation}\label{NSAM:loss}
	L_g = -[y \cdot log (\hat{P}(a|s)) + (1 - y) \cdot log(1-\hat{P}(a|s))]
\end{equation}     
The intuition of this loss is straightforward: at each $s$ it encourages the PSDD to generate higher probability to actions that are explorable (when $y=1$), and generate lower probability to those that are not explorable (when $y=0$).

\section{Combining symbolic reasoning with gradient-based DRL}
\label{NSAM:sec4}


Through the training of the gating function defined in Equation (\ref{NSAM:loss}), the PSDD in Figure \ref{fig:Framework} can predict, for a given DRL state, a distribution over the symbolic model $\bm{m}$ for atomic propositions in $\mathcal{P}$. This distribution then can be used to evaluate the truth values of the preconditions in $\mathcal{AP}$ and to reason about the explorability of actions. However, directly applying symbolic logical formula of preconditions to take actions results in non-differentiable decision-making \cite{sg2_1}, which prevents gradient flow during policy optimization. This raises a key challenge on integrating symbolic reasoning with gradient-based DRL training in a way that preserves differentiability, i.e., problem (P4) in Section \ref{NSAM:probsetting}.

To address this issue, we employ the PSDD to perform maximum a posteriori (MAP) query \cite{PCbooks}, obtaining the most likely model $\hat{\bm{m}}$ for the current state. Based on $\hat{\bm{m}}$ and the precondition $\varphi$ of each action $a$, we re-normalize the action probabilities from a policy network. In this way, the learned symbolic representation from the PSDD can be used to constrain action selection, while the underlying policy network still provides a probability distribution that can be updated through gradient-based optimization.

Concretely, before the DRL agent makes a decision, we first use the PSDD to obtain the most likely model describing the state:
\begin{equation} \label{NSAM:equ_argmax}
	\hat{\bm{m}} = argmax_{\bm{m}}  Pr(\bm{m}|\bm{\Theta}=g(s), \bm{m}\models\phi)
\end{equation}
Importantly, the argmax operation on the PSDD does not require enumerating all possible $\bm{m}$. Instead, it can be computed in linear time with respect to the PSDD size by exploiting its structural properties on decomposability and Determinism (see \cite{psdd_infer1}). This linear-time inference makes PSDDs particularly attractive for DRL, where efficient evaluation of candidate actions are essential \cite{anokhinhandling}.

After obtaining the symbolic model of the state, we renormalize the probability of each action $a$ according to its precondition $\varphi$:
\begin{equation} \label{ASG:equ_actionmask}
	\pi^{+}(s,a,\phi) =
	\frac{\pi(s,a)\cdot C_{\varphi}(\hat{\bm{m}})}
	{\sum_{a' \in \mathcal{A}} \pi(s,a')\cdot C_{\varphi'}(\hat{\bm{m}})}
\end{equation}
where $\pi(s,a)$ denotes the probability of action $a$ at state $s$ predicted by the policy network, $C_{\varphi}(\hat{\bm{m}})$ is the evaluation of the SDD encoding from $\varphi$ under the model $\hat{\bm{m}}$, and $\varphi'$ is the precondition of action $a'$. The input of Equation~(\ref{ASG:equ_actionmask}) explicitly includes $\phi$, as $\phi$ is required for evaluating the model $\hat{\bm{m}}$ in Equation~(\ref{NSAM:equ_argmax}). Intuitively, $C_{\varphi}(\hat{\bm{m}})$ acts as a symbolic mask. It equals to 1 if $\hat{\bm{m}} \models \varphi$ (i.e., the precondition is satisfied) and 0 otherwise. As a result, actions whose preconditions are violated are excluded from selection, while the probabilities of the remaining actions are renormalized as a new distribution. It is important to note that during the execution, we use the PSDD (trained by $y$ in Equation~(\ref{NSAM:hatp}) and (\ref{NSAM:loss}) ) to infer the most probable symbolic model of the current state (in Equation~(\ref{NSAM:equ_argmax})), and therefore can formally verify whether each action's precondition is satisfied with this symbolic model (happened in $C_{\varphi}$ in Equation~(\ref{ASG:equ_actionmask})).

According to prior work, such $0$-$1$ masking and renormalization still yield a valid policy gradient, thereby preserving the theoretical guarantees of policy optimization \cite{actionmasking_vPG}. In practice, we optimize the masked policy $\pi^{+}$ using the Proximal Policy Optimization (PPO) objective \cite{schulman2017ppo}. Concretely, the loss is:
\begin{equation} \label{NSAM:equ_ppo}
	\mathcal{L}_{\text{PPO}}(\pi^{+}) 
	= \mathbb{E}_{t}\!\left[
	\min\!\Big(
	\mathfrak{r}_t(\pi^{+}) \,\hat{A}_t,
	\text{clip}(\mathfrak{r}_t(\pi^{+}), 1-\epsilon, 1+\epsilon)\,\hat{A}_t
	\Big)
	\right]
\end{equation}
where $\mathfrak{r}_t(\pi^{+})$ denotes the probability ratio between the new and old masked policies, `$\text{clip}$' is the clip function and $\hat{A}_t$ is the advantage estimate \cite{schulman2017ppo}. In this way, the masked policy can be trained with PPO to effectively exploit symbolic action preconditions, leading to safer and more sample-efficient learning.


\section{End-to-end training framework}
\label{NSAM:sec_meth05}

After deriving the gating function loss of PSDD in Equation (\ref{NSAM:loss}) and the DRL policy loss in Equation (\ref{NSAM:equ_ppo}), we now introduce an end-to-end training framework that combines the two components.

Before presenting the training procedure, we first summarize how the agent makes decisions, as illustrated in Figure~4. At each time step, the state $s$ is first input into the symbolic grounding module, whose internal structure is shown in Figure~3. Within this module, the PSDD produces the most probable symbolic description of the state, i.e., a model $\hat{\bm{m}}$, according to Equation (4). The agent then leverages the preconditions in $\mathcal{AP}$ (following Equation (\ref{ASG:equ_actionmask})) to mask the action distribution from policy network, and samples an action from the renormalized distribution to interact with the environment.

\begin{figure}[t]
	\centering
	\includegraphics[width=0.7\linewidth]{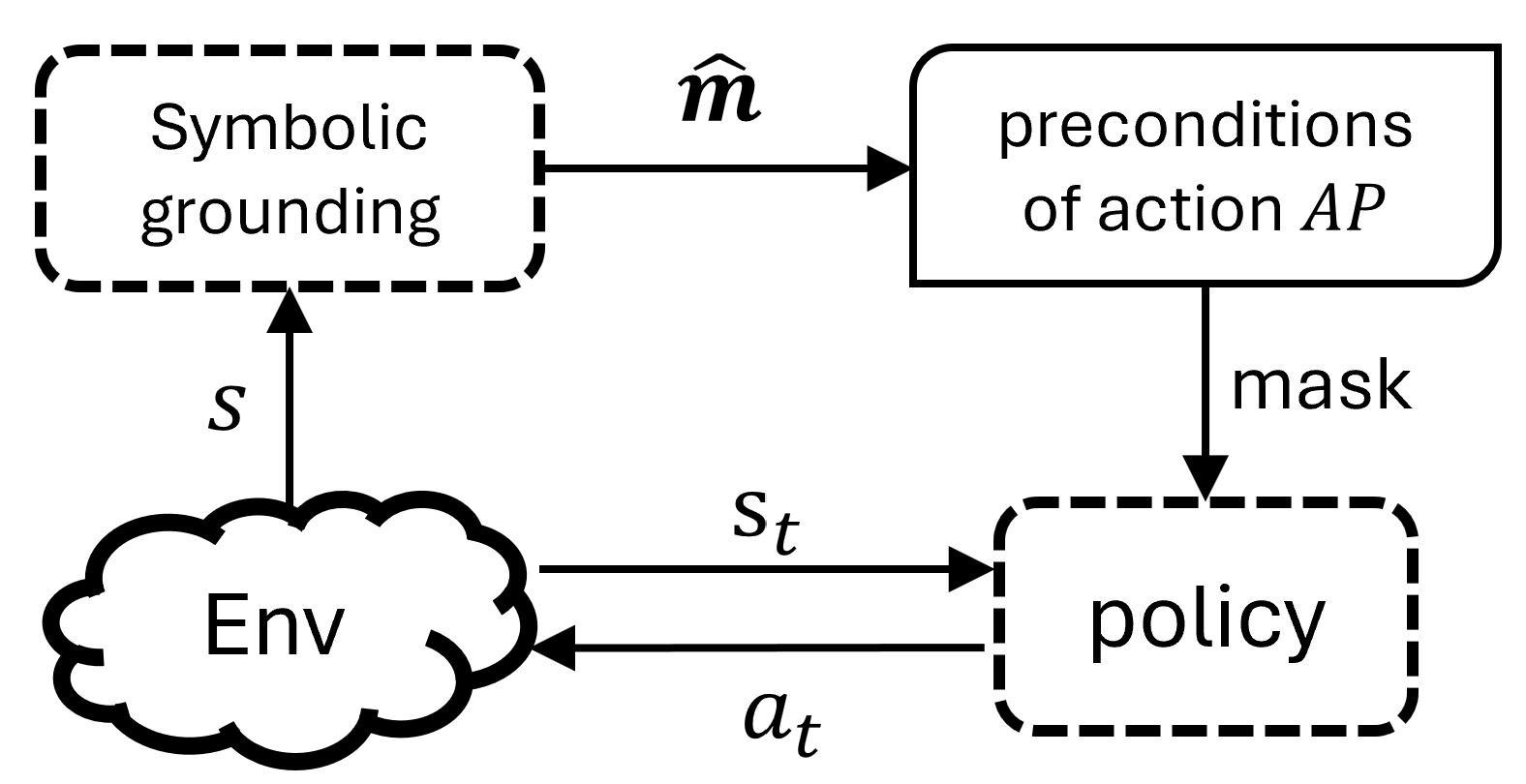}
	\caption{An illustration of the decision process of our agent, where the symbolic grounding module is as in Figure \ref{fig:Framework} and $\hat{\bm{m}}$ is calculated via the PSDD by Equation (\ref{NSAM:equ_argmax}).}
	\vspace{-0.4cm}
	\label{fig:logo}
	\Description{An illustration of the decision process of our agent.}
\end{figure}

\begin{algorithm}[t] \small
	\caption{Training framework.}
	\label{alg:framework}
	\begin{algorithmic}[1]
		\State Compile $\phi$ as SDD to obtain structure of PSDD 
		\State Initialize gating network $g$ according to the structure of PSDD
		\State Initialize policy network $\pi$, total step $T \leftarrow 0$
		\State Initialize a data buffer $\mathcal{D}$ for learning PSDD
		\For{$Episode = 1 \to M$}
		\State Reset $Env$ and get $s$
		\While{not terminal}
		\State Calculate action distribution before masking $\pi(s,a)$
		\State Calculate $\Theta=g(s)$ and assign parameter $\Theta$ to PSDD
		\State Calculate $\hat{\bm{m}}$ in Equation (\ref{NSAM:equ_argmax})
		\State Calculate action distribution after masking $\pi^+(s,a,\phi)$
		\State Sample an action $a$ from $\pi^+(s,a,\phi)$
		\State Execute $a$ and get $r$, $s'$ from $Env$ 
		\State Obtain the truth-value $y$ according to Equ. (\ref{NSAM:DATA}) 
		\State Store $(s, a, \varphi)$ into $\mathcal{D}$
		\If{terminal}
		\State Update policy $\pi^+(s,a,\phi)$ using the trajectory of this episode with Equation (\ref{NSAM:equ_ppo})
		\EndIf
		\If{$(T+1) \;\%\; freq_{g} == 0$}
		\State Sample batches from $\mathcal{D}$
		\State Update gating function $g$ with Equation (\ref{NSAM:loss})
		\EndIf
		\State $s \leftarrow s'$, $T \leftarrow T+1$
		\EndWhile
		\EndFor
	\end{algorithmic}
\end{algorithm}

To achieve end-to-end training, we propose Algorithm~1. The key idea for this training framework is to periodically update the gating function of the PSDD during the agent’s interaction with the environment, while simultaneously training the policy network under the guidance of action masks. As the RL agent explores, it continuously generates minimally supervised feedback for the PSDD via $\Gamma_{\phi}$, thereby improving the quality of the learned action masks. In turn, the improved action masking reduces infeasible actions and guides the agent toward higher rewards and more informative trajectories, which accelerates policy learning.

Concretely, before the start of training, the domain constrain $\phi$ is compiled into an SDD in Line 1, which determines both the structure of the PSDD and the output dimensionality of the gating function. Lines 2$\sim$4 initialize the gating function, the RL policy network, and a replay buffer $\mathcal{D}$ that stores minimally supervised feedback for PSDD training. In Lines 5$\sim$25, the agent interacts with the environment and jointly learns the gating function and policy network. At each step (lines 8$\sim$11), the agent computes the masked action distribution based on the current gating function and policy network, which is crucial to minimizing the selection of infeasible actions during training. At the end of each episode (lines 16$\sim$18), the policy network is updated using the trajectory of this episode. In this process, the gating function is kept frozen. In addition, the gating function is periodically updated (lines 19$\sim$22) with frequency $freq_{g}$. This periodically update enables the PSDD to provide increasingly accurate action masks in subsequent interactions, which simultaneously improves policy optimization and reduces constraint violations.

\section{Related work}

\textbf{Symbolic grounding in neuro-symbolic learning.} In the literature on neuro-symbolic systems \cite{NSsys, NSsys2}, symbol grounding refers to learning a mapping from raw data to symbolic representations, which is considered as a key challenge for integrating neural networks with symbolic knowledge \cite{neuroRM}. Various approaches have been proposed to address this challenge in deep supervised learning. \cite{sg1_1, sg1_2, sg1_3} leverage logic abduction or consistency checking to periodically correct the output symbolic representations. To achieve end-to-end differentiable training, the most common methods are embedding symbolic knowledge into neural networks through differentiable relaxations of logic operators, such as fuzzy logic \cite{sg2_2} or Logic Tensor Networks (LTNs) \cite{sg2_1}. These methods approximate logical operators with smooth functions, allowing symbolic supervision to be incorporated into gradient-based optimization \cite{sg2_2, sg2_3, sg2_4, neuroRM}. More recently, advances in probabilistic circuits \cite{psdd_infer1, PCbooks} give rise to efficient methods that embed symbolic knowledge via differentiable probabilistic representations, such as PSDD \cite{kisa2014probabilistic}. In these methods, symbolic knowledge is first compiled into SDD \cite{sdd} to initialize the structure, after which a neural network is used to learn the parameters for predicting symbolic outputs \cite{ahmed2022semantic}. This class of approaches has been successfully applied to structured output prediction tasks, including multi-label classification \cite{ahmed2022semantic} and routing \cite{psdd_infer2}. 

Symbolic grounding is also crucial in DRL. NRM \cite{neuroRM} learn to capture the symbolic structure of reward functions in non-Markovian reward settings. In contrast, our approach learns symbolic properties of states to constrain actions under Markovian reward settings. KCAC \cite{KCAC} has extended PSDDs to MDP with combinatorial action spaces, where symbolic knowledge is used to constrain action composition. Our work also uses PSDDs but differs from KCAC. We use preconditions of actions as symbolic knowledge to determine the explorability of each individual action in a standard DRL setting, whereas KCAC incorporates knowledge about valid combinations of actions in a DRL setting with combinatorial action spaces. 


\textbf{Action masking.} In DRL, action masking refers to masking out invalid actions during training to sample actions from a valid set \cite{actionmasking_vPG}. Empirical studies in early real-world applications show that masking invalid actions can significantly improve sample efficiency of DRL \cite{actionmasking_app1, actionmasking_app2, actionmasking_app3, actionmasking_app4, actionmasking_app5, actionmasking_app6}. Following the systematic discussion of action masking in DRL \cite{actionmasking_review}, \cite{actionmasking_onoffpolicy} investigates the impact of action masking on both on-policy and off-policy algorithms. Works such as \cite{actionmasking_continous, actionmasking_continous2} extend action masking to continuous action spaces. \cite{actionmasking_vPG} proves that binary action masking have Valid policy gradients during learning. In contrast to these approaches, our method does not assume that the set of invalid actions is predefined by the environment. Instead, we learn the set of invalid actions in each state for DRL using action precondition knowledge.

Another line of work employs a logical system (e.g., linear temporal logic \cite{LTL1985}) to restrict the agent’s actions \cite{shielding1, shielding2, shielding3, PPAM}. These approaches require a predefined symbol grounding function to map states into its symbolic representations, whereas our method learn such function (via PSDD) from data. PLPG \cite{PLPG} learns the probability of applying shielding with action constraints formulated in probabilistic logics. By contrast, our preconditions are hard constraints expressed in propositional logic: if the precondition of an action is evaluated to be false, the action is strictly not explorable.

\textbf{Cost-based safe reinforcement learning.} In addition to action masking, a complementary approach is to jointly optimize rewards and a safety-wise cost function to improve RL safety. In these cost-based settings, a policy is considered safe if its expected cumulative cost remains below a pre-specified threshold \cite{safeexp1, PPOlarg}. A representative foundation of such cost-based approach is the constrained Markov decision process (CMDP) framework \cite{CMDP1994}, which aims to maximize expected reward while ensuring costs below a threshold. Subsequent works often adopt Lagrangian relaxation to incorporate constraints into the optimization objective \cite{ppo-larg1, ppo-larg2, ppo-larg3, ppo-larg4, PPOlarg}. However, these methods often suffer from unsafe behaviors in the early stages of training \cite{highvoil1}. To address such issues, safe exploration approaches emphasize to control the cost during exploration in unknown environments \cite{shielding1, shielding2}. Recently, SaGui \cite{safeexp1} employed imitation learning and policy distillation to enable agents to acquire safe behaviors from a teacher agent during early training. RC-PPO \cite{RCPPO} augmented unsafe states to allow the agent to anticipate potential future losses. While constraints can in principle be reformulated as cost functions, our approach does not rely on cost-based optimization. Instead, we directly exploit them to learn masks to avoid the violation of actions constraints.

\section{Experiment}

The experimental design aims to answer the following questions:

Q1: Without predefined symbolic grounding, can NSAM leverage symbolic knowledge to improve the sample efficiency of DRL?

Q2: By jointly learning symbolic grounding and masking strategies, can NSAM significantly reduce constraint violations during exploration, thereby enhancing safety?

Q3: In NSAM, is symbolic grounding with PSDDs more effective than replacing it with a module based on standard neural network?

Q4: In what ways does symbolic knowledge contribute to the learning process of NSAM?

\subsection{Environments}

\begin{figure}[t]
	\centering
	\subfigure[Sudoku]{\includegraphics[width=2cm]{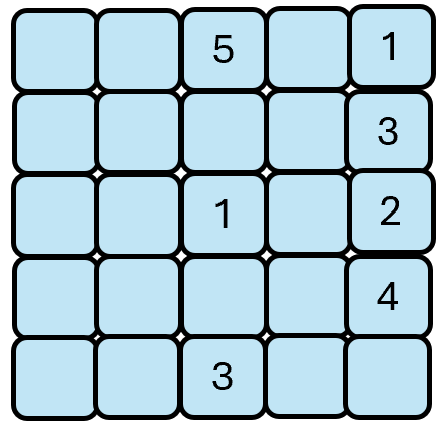}\label{fig:Sudoku}}
	\subfigure[N-queens]{\includegraphics[width=1.9cm]{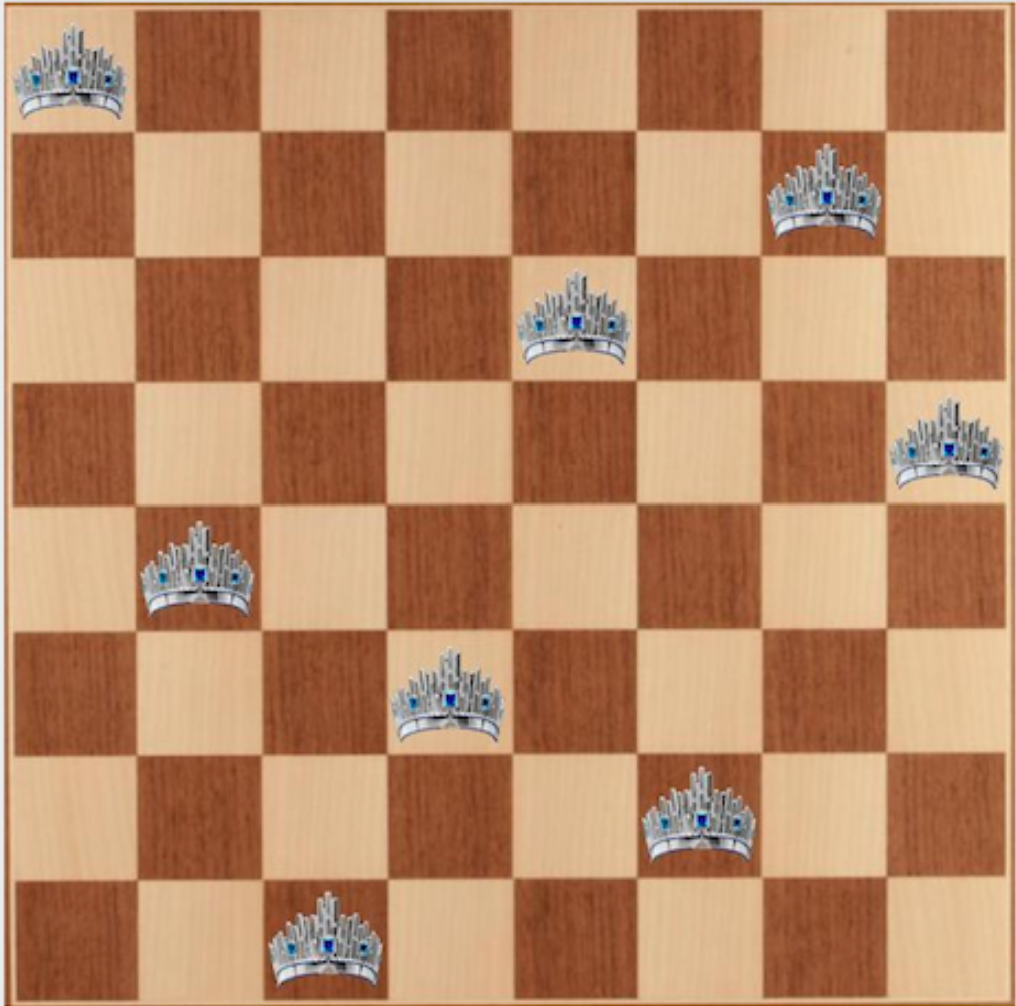}\label{fig:queens}}
	\subfigure[Graph coloring]{\includegraphics[width=2cm]{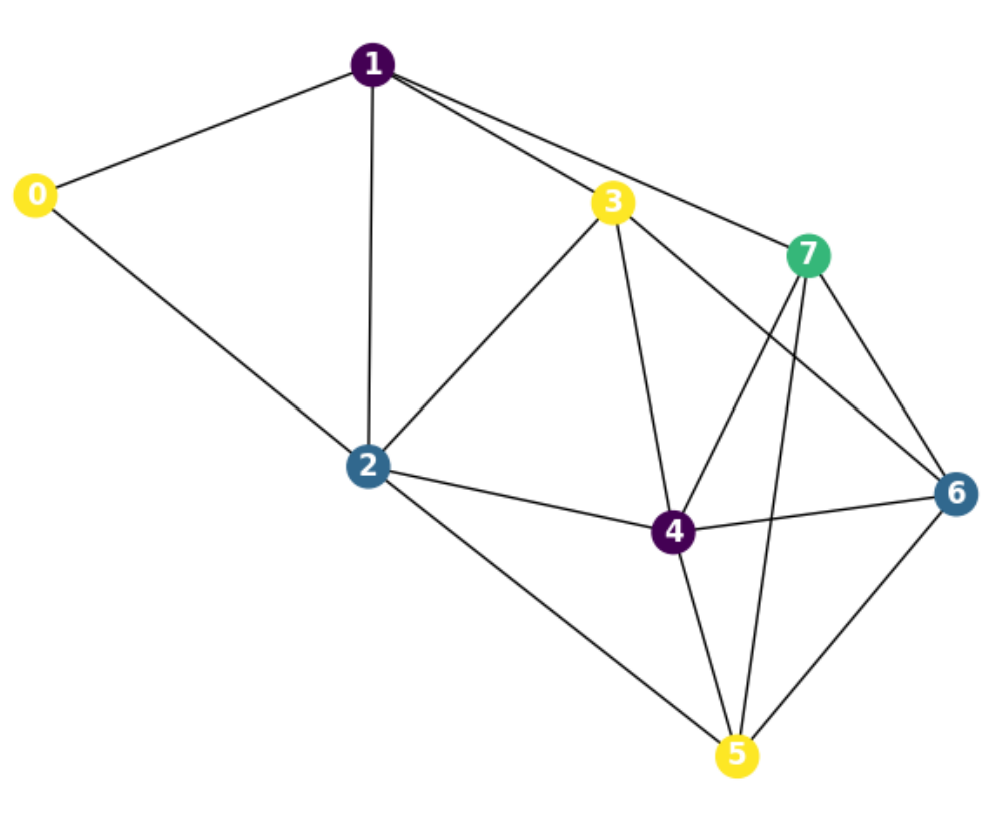}\label{fig:coloring}}
	\subfigure[Visual Sudoku]{\includegraphics[width=2cm]{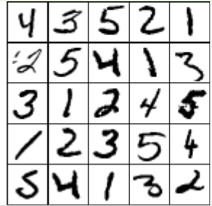}\label{fig:Sudokuv}}
	\vspace{-0.2cm}
	\caption{Four tasks with logical constraints}
	\label{fig:env}
\end{figure}

We evaluate ASG on four highly challenging reinforcement learning domains with logical constraints as shown in Figure \ref{fig:env}. Across all these environments, agents receive inputs in the form of unknown representation such as vectors or images.

\begin{figure*}[t]
	\centering
	\subfigure[Sudoku 3$\times$3]{\includegraphics[width=3.2cm]{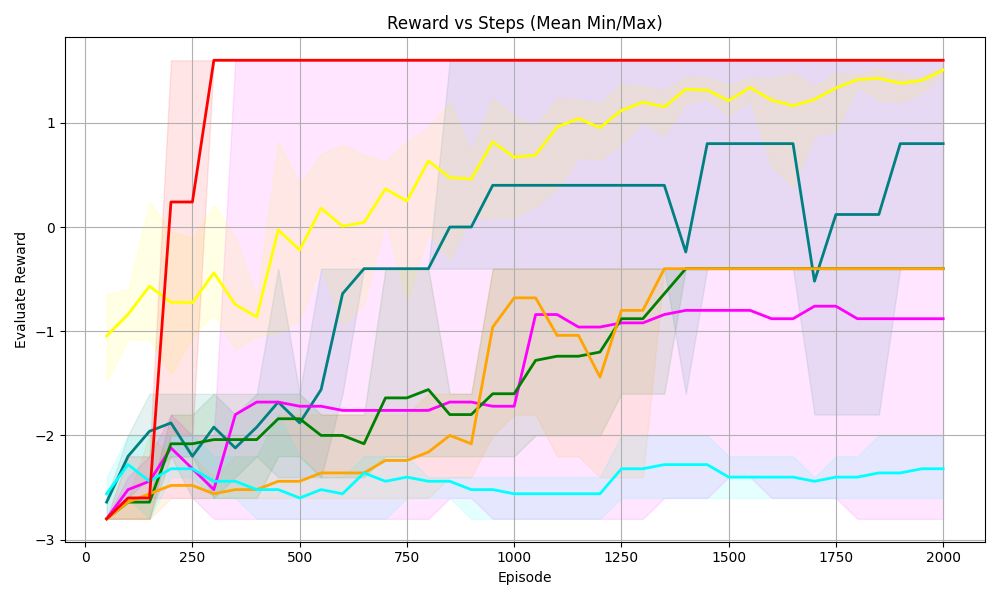}}
	\subfigure[Sudoku 4$\times$4]{\includegraphics[width=3.2cm]{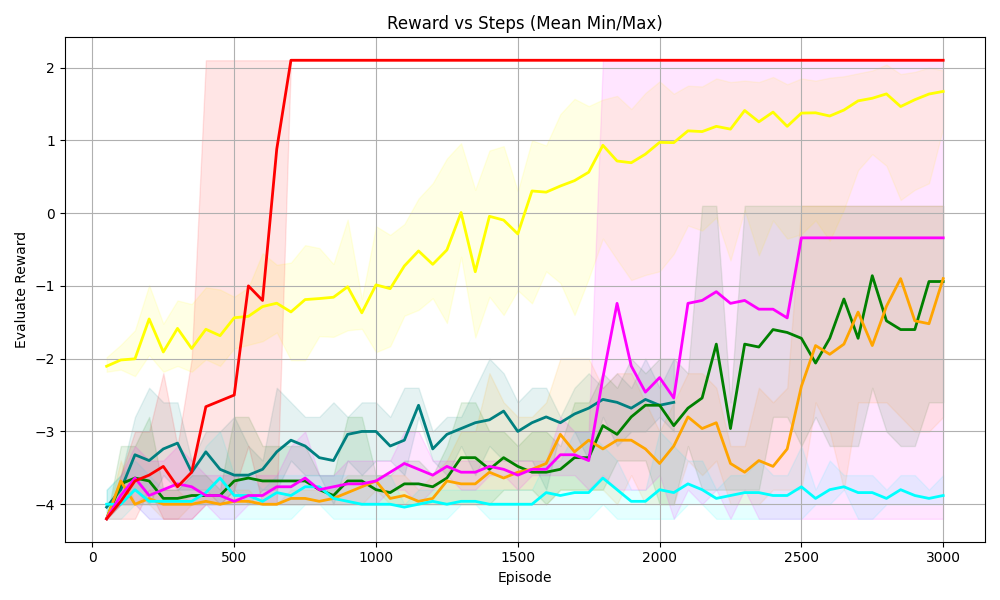}}
	\subfigure[Sudoku 5$\times$5]{\includegraphics[width=3.2cm]{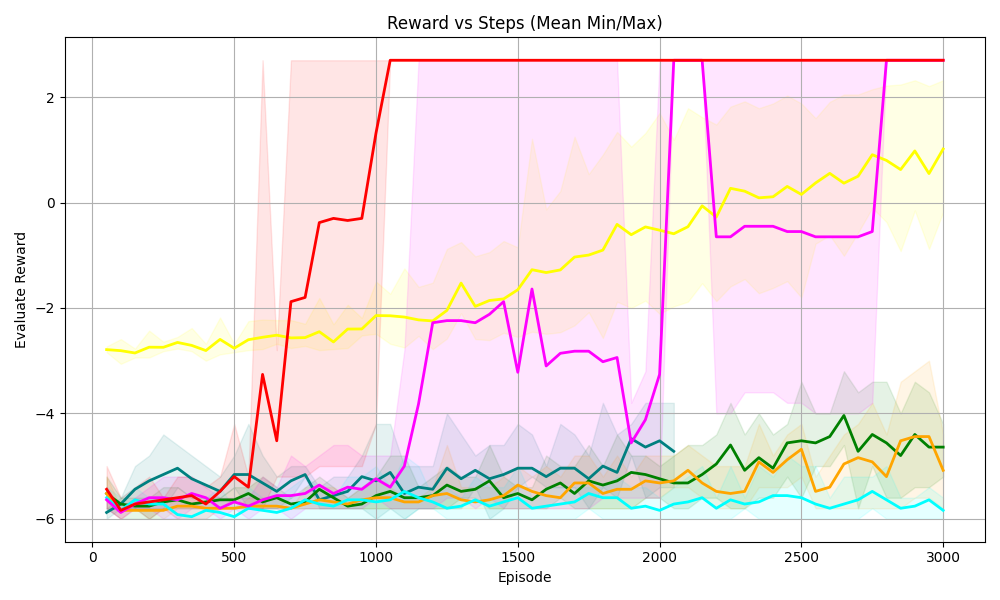}}
	\subfigure[4 Queens]{\includegraphics[width=3.2cm]{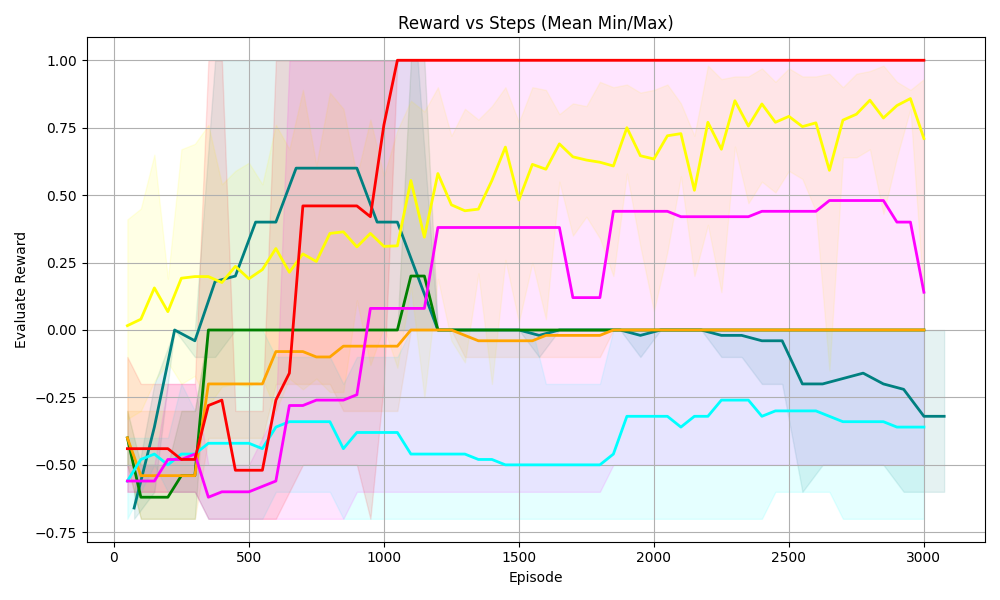}}
	\subfigure[6 Queens]{\includegraphics[width=3.2cm]{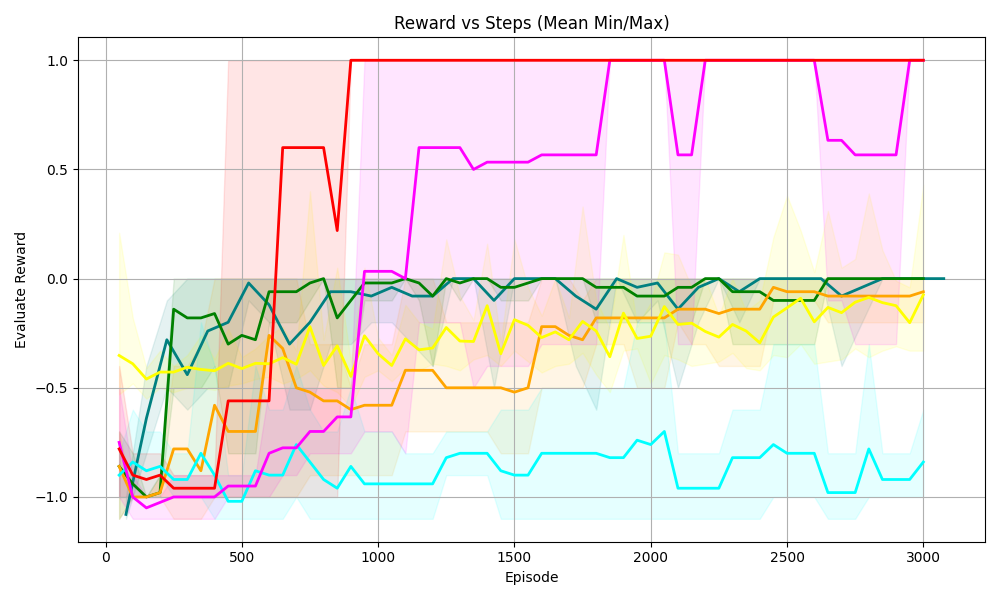}}
	\vspace{-0.2cm}
	
	\subfigure[8 Queens]{\includegraphics[width=3.2cm]{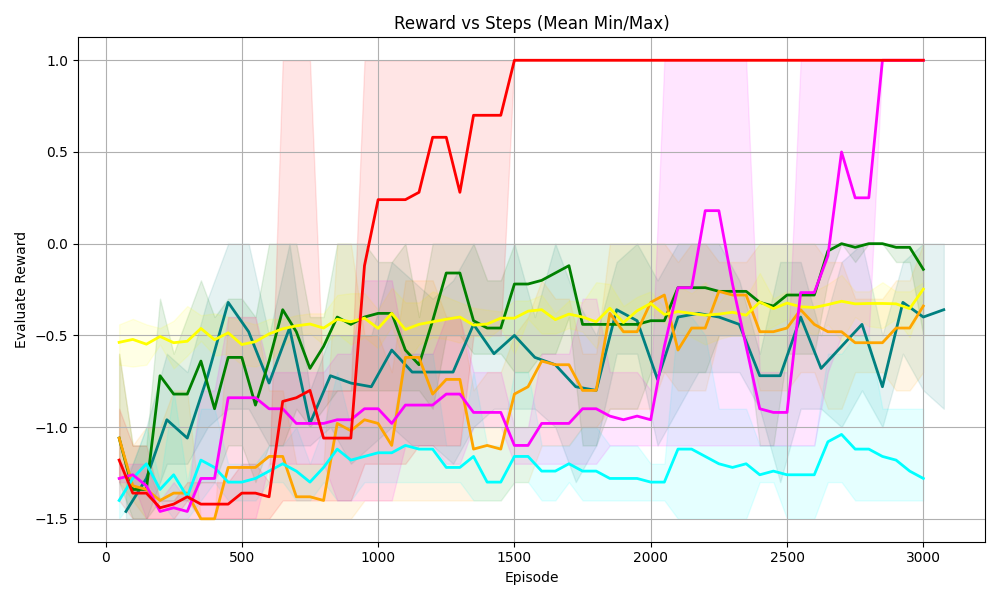}}
	\subfigure[10 Queens]{\includegraphics[width=3.2cm]{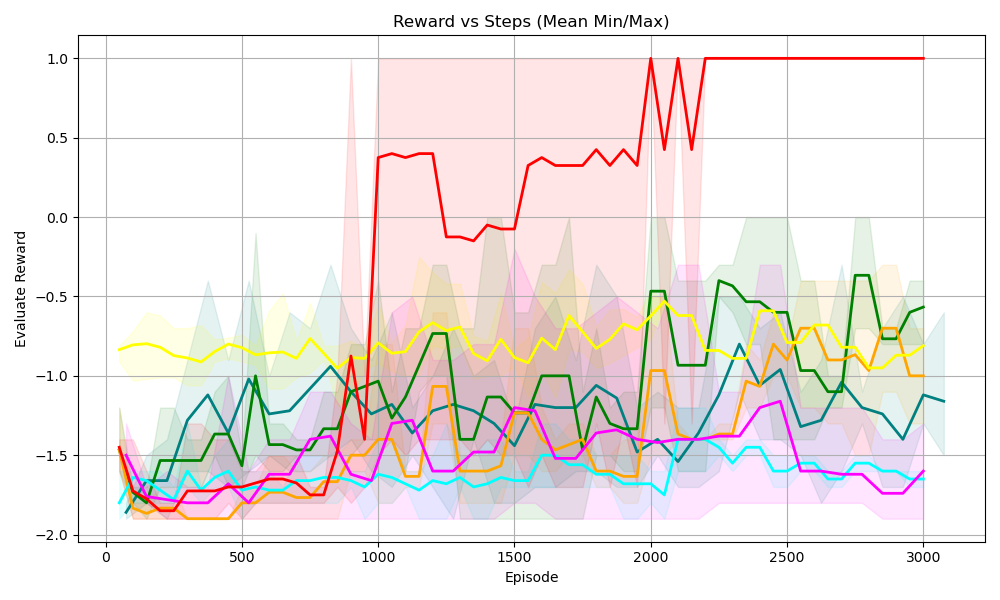}}
	\subfigure[Graph 1]{\includegraphics[width=3.2cm]{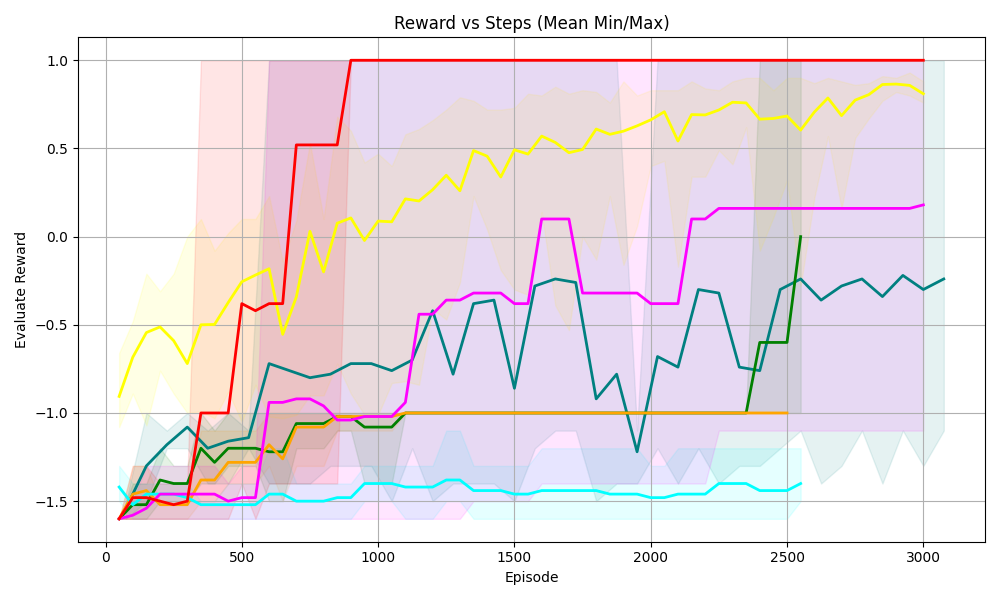}}
	\subfigure[Graph 2]{\includegraphics[width=3.2cm]{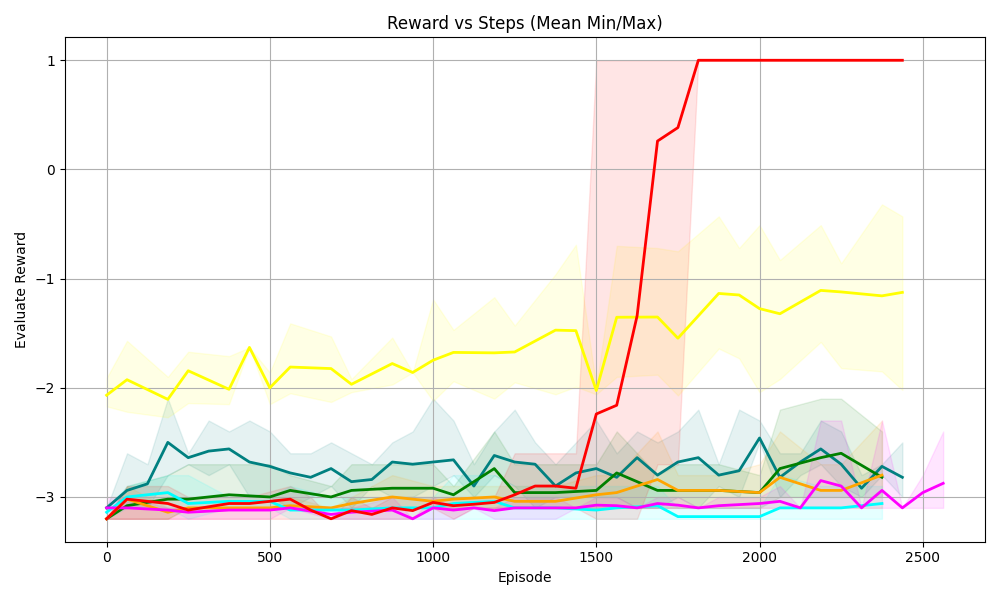}}
	\subfigure[Graph 3]{\includegraphics[width=3.2cm]{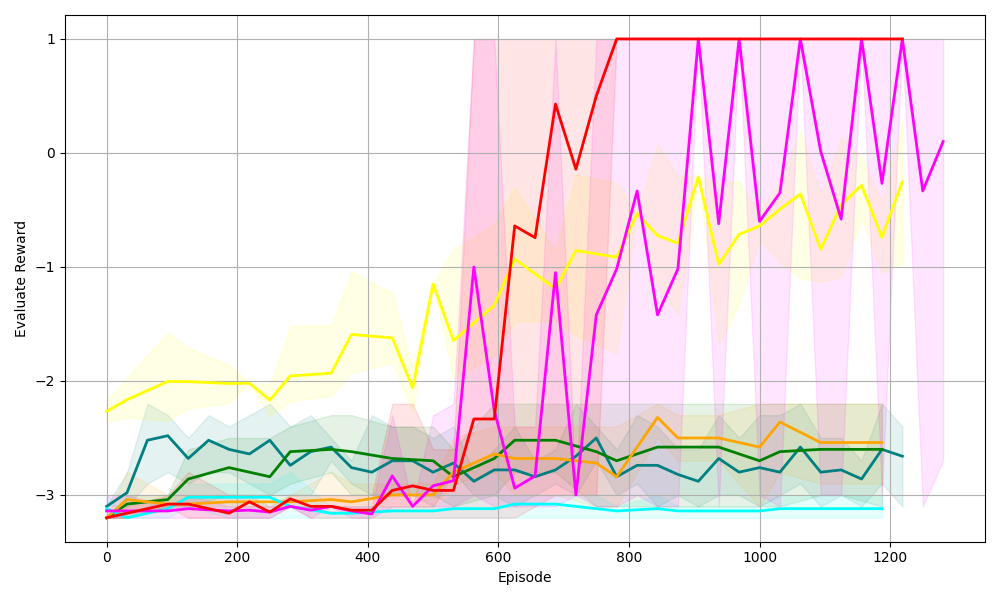}}
	\vspace{-0.2cm}
	
	\subfigure[Graph 4]{\includegraphics[width=3.2cm]{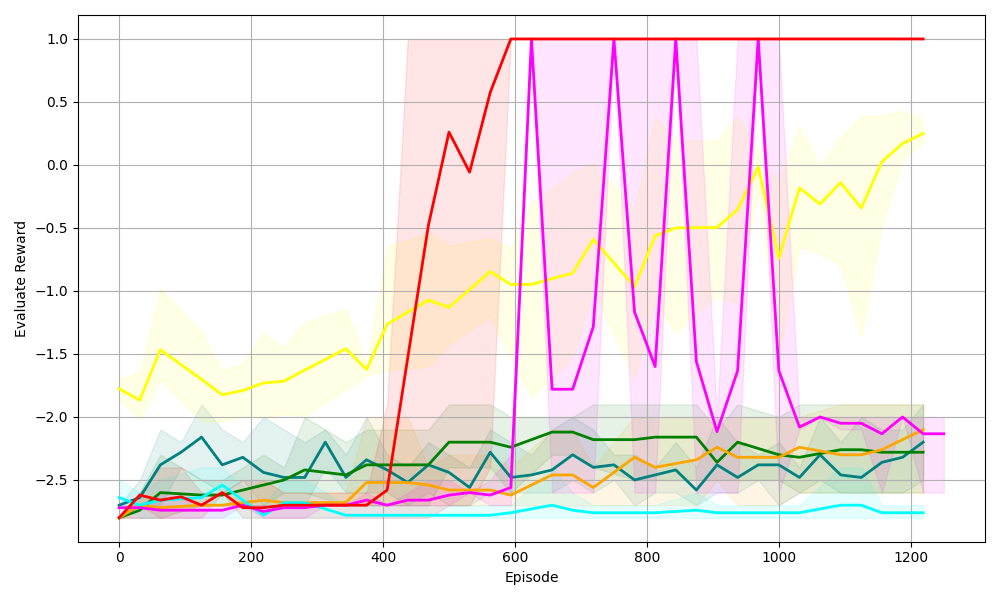}}
	\subfigure[Visual Sudoku 2$\times$2]{\includegraphics[width=3.2cm]{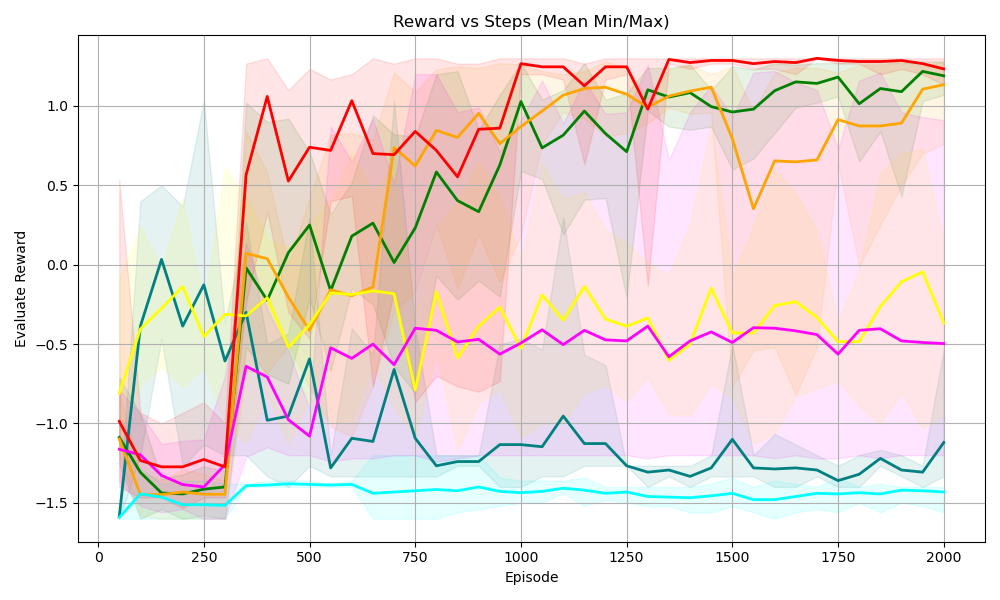}}
	\subfigure[Visual Sudoku 3$\times$3]{\includegraphics[width=3.2cm]{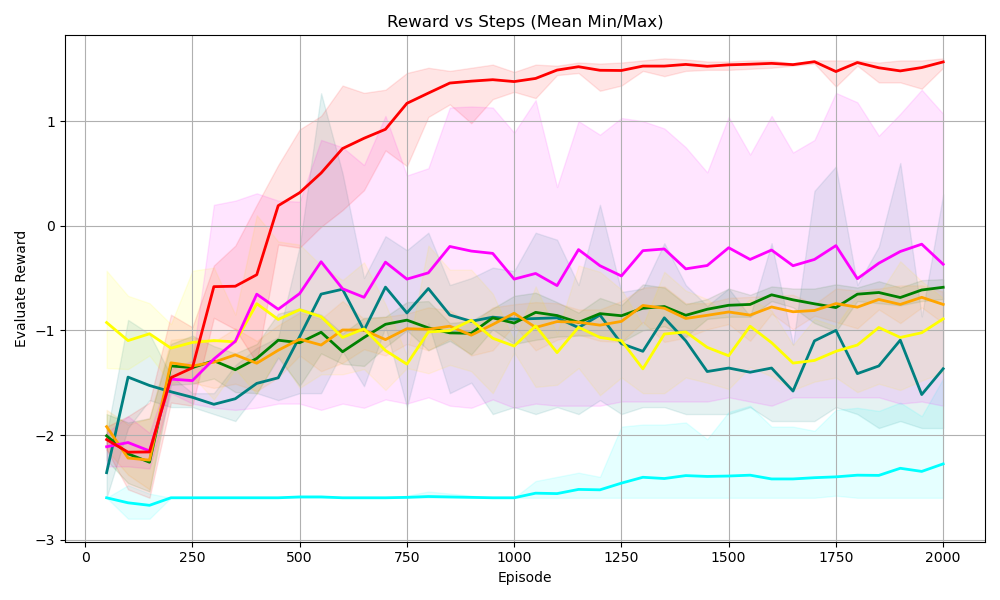}}
	\subfigure[Visual Sudoku 4$\times$4]{\includegraphics[width=3.2cm]{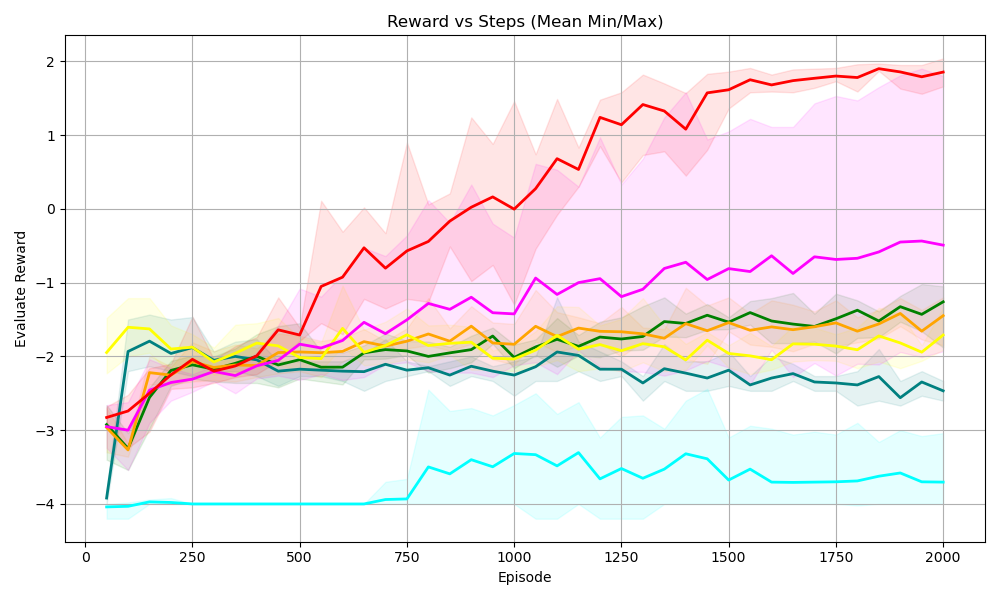}}
	\subfigure[Visual Sudoku 5$\times$5]{\includegraphics[width=3.2cm]{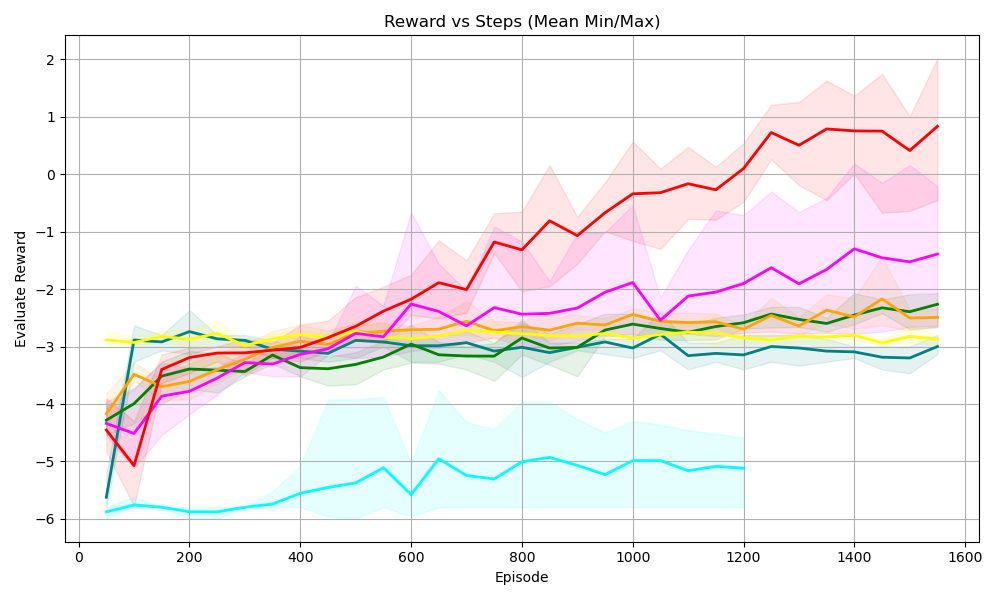}}
	\vspace{-0.2cm}
	
	\subfigure{\includegraphics[width=14cm]{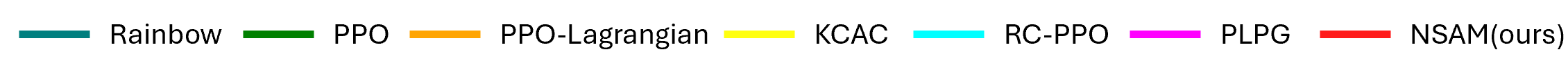}}
	\vspace{-0.4cm}
	\caption{Comparsion of learning curves on 4 domains. As the size and queen number increase in Sudoku and N-Queens, the learning task becomes more challenging. Graphs 1$\sim$4 denote four graph coloring tasks with different topologies.}
	\label{fig:compare}
\end{figure*}

\textbf{Sudoku.} Sudoku is a decision-making domain with logical constraints \cite{sudokuRL}. In this domain, the agent fills one cell with a number at each step until the board is complete. Action preconditions naturally arise: filling a number to a cell requires that the same number does not already exist in this row and column. Prior work \cite{sudokuRL, sudokuMLP} shows that existing DRL algorithms struggle to solve Sudoku without predefined symbolic grounding functions.



\begin{table*}[] \small
	\begin{tabular}{|ll|ll|ll|ll|ll|ll|ll|ll|}
		\hline
		\multicolumn{2}{|l|}{\multirow{2}{*}{}}                                                                 & \multicolumn{2}{c|}{Rainbow}               & \multicolumn{2}{c|}{PPO}                   & \multicolumn{2}{c|}{PPO-lagrangian.}             & \multicolumn{2}{c|}{KCAC}                  & \multicolumn{2}{c|}{RC-PPO}        & \multicolumn{2}{c|}{PLPG}                  & \multicolumn{2}{c|}{NSAM(ours)}                          \\ \cline{3-16} 
		\multicolumn{2}{|l|}{}                                                                                  & \multicolumn{1}{l|}{Rew.}         & Viol.   & \multicolumn{1}{l|}{Rew.}         & Viol.   & \multicolumn{1}{l|}{Rew.}         & Viol.   & \multicolumn{1}{l|}{Rew.}         & Viol.   & \multicolumn{1}{l|}{Rew.} & Viol.   & \multicolumn{1}{l|}{Rew.}         & Viol.   & \multicolumn{1}{l|}{Rew.}         & Viol.           \\ \hline
		\multicolumn{1}{|l|}{\multirow{4}{*}{Sudoku}}                                                    & 2×2  & \multicolumn{1}{l|}{\textbf{1.3}} & 36.2\% & \multicolumn{1}{l|}{\textbf{1.3}} & 14.9\% & \multicolumn{1}{l|}{\textbf{1.3}} & 0.7\%  & \multicolumn{1}{l|}{\textbf{1.3}} & 9.2\%  & \multicolumn{1}{l|}{0.3}  & 11.3\% & \multicolumn{1}{l|}{-0.8}         & 5.9\%  & \multicolumn{1}{l|}{\textbf{1.3}} & \textbf{0.1\%} \\ \cline{3-16} 
		\multicolumn{1}{|l|}{}                                                                           & 3×3  & \multicolumn{1}{l|}{0.8}          & 88.2\% & \multicolumn{1}{l|}{-0.4}         & 99.6\% & \multicolumn{1}{l|}{-0.4}         & 99.9\% & \multicolumn{1}{l|}{1.5}          & 57.1\% & \multicolumn{1}{l|}{-2.3} & 99.0\% & \multicolumn{1}{l|}{-0.9}         & 8.9\%  & \multicolumn{1}{l|}{\textbf{1.6}} & \textbf{0.3\%} \\ \cline{3-16} 
		\multicolumn{1}{|l|}{}                                                                           & 4×4  & \multicolumn{1}{l|}{-2.6}         & 100\%  & \multicolumn{1}{l|}{-2.6}         & 100\%  & \multicolumn{1}{l|}{-3.4}         & 100\%  & \multicolumn{1}{l|}{1.0}          & 91.2\% & \multicolumn{1}{l|}{-3.8} & 99.9\% & \multicolumn{1}{l|}{-2.2}         & 15.7\% & \multicolumn{1}{l|}{\textbf{2.1}} & \textbf{0.6\%} \\ \cline{3-16} 
		\multicolumn{1}{|l|}{}                                                                           & 5×5  & \multicolumn{1}{l|}{-4.5}         & 100\%  & \multicolumn{1}{l|}{-5.2}         & 100\%  & \multicolumn{1}{l|}{-5.3}         & 100\%  & \multicolumn{1}{l|}{-0.5}         & 94.9\% & \multicolumn{1}{l|}{-5.8} & 100\%  & \multicolumn{1}{l|}{-3.3}         & 18.3\% & \multicolumn{1}{l|}{\textbf{2.7}} & \textbf{4.3\%} \\ \hline
		\multicolumn{1}{|l|}{\multirow{4}{*}{N-Queens}}                                                  & N=4  & \multicolumn{1}{l|}{-0.3}         & 97.8\% & \multicolumn{1}{l|}{0.0}          & 99.8\% & \multicolumn{1}{l|}{0.0}          & 100\%  & \multicolumn{1}{l|}{0.7}          & 78.7\% & \multicolumn{1}{l|}{-0.3} & 93.6\% & \multicolumn{1}{l|}{0.1}          & 10.4\% & \multicolumn{1}{l|}{\textbf{1.0}} & \textbf{2.3\%} \\ \cline{3-16} 
		\multicolumn{1}{|l|}{}                                                                           & N=6  & \multicolumn{1}{l|}{0.0}          & 100\%  & \multicolumn{1}{l|}{0.0}          & 100\%  & \multicolumn{1}{l|}{-0.1}         & 100\%  & \multicolumn{1}{l|}{-0.1}         & 100\%  & \multicolumn{1}{l|}{-0.8} & 100\%  & \multicolumn{1}{l|}{\textbf{1.0}} & 12.1\% & \multicolumn{1}{l|}{\textbf{1.0}} & \textbf{1.3\%} \\ \cline{3-16} 
		\multicolumn{1}{|l|}{}                                                                           & N=8  & \multicolumn{1}{l|}{-0.4}         & 100\%  & \multicolumn{1}{l|}{-0.1}         & 100\%  & \multicolumn{1}{l|}{-0.3}         & 100\%  & \multicolumn{1}{l|}{-0.2}         & 100\%  & \multicolumn{1}{l|}{-1.3} & 100\%  & \multicolumn{1}{l|}{\textbf{1.0}} & 41.6\% & \multicolumn{1}{l|}{\textbf{1.0}} & \textbf{1.1\%} \\ \cline{3-16} 
		\multicolumn{1}{|l|}{}                                                                           & N=10 & \multicolumn{1}{l|}{-1.2}         & 100\%  & \multicolumn{1}{l|}{-0.6}         & 100\%  & \multicolumn{1}{l|}{-1.0}         & 100\%  & \multicolumn{1}{l|}{-0.8}         & 100\%  & \multicolumn{1}{l|}{-1.7} & 100\%  & \multicolumn{1}{l|}{-1.6}         & 98.2\% & \multicolumn{1}{l|}{\textbf{1.0}} & \textbf{1.5\%} \\ \hline
		\multicolumn{1}{|l|}{\multirow{4}{*}{\begin{tabular}[c]{@{}l@{}}Graph \\ Coloring\end{tabular}}} & G1   & \multicolumn{1}{l|}{-0.2}         & 88.7\% & \multicolumn{1}{l|}{0.0}          & 98.6\% & \multicolumn{1}{l|}{-1.0}         & 100\%  & \multicolumn{1}{l|}{0.8}          & 43.9\% & \multicolumn{1}{l|}{-1.4} & 99.1\% & \multicolumn{1}{l|}{0.2}          & 5.9\%  & \multicolumn{1}{l|}{\textbf{1.0}} & \textbf{0.7\%} \\ \cline{3-16} 
		\multicolumn{1}{|l|}{}                                                                           & G2   & \multicolumn{1}{l|}{-2.8}         & 100\%  & \multicolumn{1}{l|}{-2.8}         & 100\%  & \multicolumn{1}{l|}{-2.8}         & 100\%  & \multicolumn{1}{l|}{-1.1}         & 98.7\% & \multicolumn{1}{l|}{-3.1} & 98.7\% & \multicolumn{1}{l|}{-2.8}         & 18.7\% & \multicolumn{1}{l|}{\textbf{1.0}} & \textbf{0.7\%} \\ \cline{3-16} 
		\multicolumn{1}{|l|}{}                                                                           & G3   & \multicolumn{1}{l|}{-2.7}         & 100\%  & \multicolumn{1}{l|}{-2.6}         & 100\%  & \multicolumn{1}{l|}{-2.5}         & 100\%  & \multicolumn{1}{l|}{-0.3}         & 90.0\% & \multicolumn{1}{l|}{-3.1} & 98.9\% & \multicolumn{1}{l|}{0.1}          & 7.5\%  & \multicolumn{1}{l|}{\textbf{1.0}} & \textbf{0.4\%} \\ \cline{3-16} 
		\multicolumn{1}{|l|}{}                                                                           & G4   & \multicolumn{1}{l|}{-2.2}         & 100\%  & \multicolumn{1}{l|}{-2.3}         & 100\%  & \multicolumn{1}{l|}{-2.1}         & 100\%  & \multicolumn{1}{l|}{0.3}          & 85.4\% & \multicolumn{1}{l|}{-2.8} & 75.8\% & \multicolumn{1}{l|}{-2.1}         & 11.2\% & \multicolumn{1}{l|}{\textbf{1.0}} & \textbf{0.2\%} \\ \hline
		\multicolumn{1}{|l|}{\multirow{4}{*}{\begin{tabular}[c]{@{}l@{}}Visual\\ Sudoku\end{tabular}}}   & 2×2  & \multicolumn{1}{l|}{-1.1}         & 61.8\% & \multicolumn{1}{l|}{1.2}          & 60.0\% & \multicolumn{1}{l|}{1.1}          & 25.0\% & \multicolumn{1}{l|}{-0.4}         & 32.2\% & \multicolumn{1}{l|}{-1.4} & 68.3\% & \multicolumn{1}{l|}{-0.5}         & 16.1\% & \multicolumn{1}{l|}{\textbf{1.2}} & \textbf{6.1\%} \\ \cline{3-16} 
		\multicolumn{1}{|l|}{}                                                                           & 3×3  & \multicolumn{1}{l|}{-1.4}         & 96.0\% & \multicolumn{1}{l|}{-0.6}         & 99.6\% & \multicolumn{1}{l|}{-0.8}         & 100\%  & \multicolumn{1}{l|}{-0.9}         & 40.3\% & \multicolumn{1}{l|}{-2.3} & 95.4\% & \multicolumn{1}{l|}{-0.4}         & 34.6\% & \multicolumn{1}{l|}{\textbf{1.5}} & \textbf{1.0\%} \\ \cline{3-16} 
		\multicolumn{1}{|l|}{}                                                                           & 4×4  & \multicolumn{1}{l|}{-2.5}         & 100\%  & \multicolumn{1}{l|}{-1.3}         & 100\%  & \multicolumn{1}{l|}{-1.4}         & 100\%  & \multicolumn{1}{l|}{-1.7}         & 39.4\% & \multicolumn{1}{l|}{-3.7} & 96.3\% & \multicolumn{1}{l|}{-0.5}         & 38.2\% & \multicolumn{1}{l|}{\textbf{1.9}} & \textbf{0.6\%} \\ \cline{3-16} 
		\multicolumn{1}{|l|}{}                                                                           & 5×5  & \multicolumn{1}{l|}{-3.0}         & 100\%  & \multicolumn{1}{l|}{-2.2}         & 100\%  & \multicolumn{1}{l|}{-2.5}         & 100\%  & \multicolumn{1}{l|}{-2.9}         & 88.5\% & \multicolumn{1}{l|}{-5.1} & 99.7\% & \multicolumn{1}{l|}{-1.4}         & 53.7\% & \multicolumn{1}{l|}{\textbf{0.8}} & \textbf{2.5\%} \\ \hline 
	\end{tabular} 
	\vspace{0.3cm}
	\caption{Comaprison of final reward (Rew.) and violation (Viol.) rate during training.}
\end{table*}

\textbf{N-Queens.} The N-Queens problem requires placing $N$ queens on a chessboard so that no two attack each other, making it a classic domain with logical constraints \cite{nqueens}. The agent places one queen per step on the chessboard until all $N$ queens are placed safely. This task fits naturally within our extended MDP framework, where action preconditions arise: a queen can be placed at a position if and only if no already-placed queen can attack it.


\textbf{Graph Coloring.} Graph coloring is a NP-hard problem and serves as a reinforcement learning domain with logical constraints \cite{Graphc}. The agent sequentially colors the nodes of an undirected graph with a limited set of colors. Action preconditions arise naturally: a node may be colored with a given color if and only if none of its neighbors have already been assigned that color.

\textbf{Visual Sudoku.} Visual Sudoku follows the same rules as standard Sudoku but uses image-based digit representations instead of vector inputs. Following prior visual Sudoku benchmark \cite{SudokuV_SATNET}, we generate boards by randomly sampling digits from the MNIST dataset \cite{mnist}. Visual Sudoku 5×5 is with a 140×140-dimensional state space and a 125-dimensional action space. In addition, the corresponding PSDD contains 125 atomic propositions and 782 clauses as constrains. This environment poses an additional challenge, as the digits are high-dimensional and uncertain representations, which increases the difficulty of symbolic grounding.


\subsection{Hyperparameters}
We run NSAM on a 2.60 GHz AMD Rome 7H12 CPU and an NVIDIA GeForce RTX 3070 GPU. For the policy and value functions, NSAM uses three-layer fully connected networks with 64 neurons per layer, while the gating function (Figure 3) uses a three-layer fully connected network with 128 neurons per layer. In the Visual Sudoku environment with image-based inputs, the policy and value functions are equipped with a convolutional encoder consisting of two convolutional layers (kernel sizes of 5 and 3, stride 2), followed by ReLU activations. The encoder output is then connected to a two-layer fully connected network with 256 neurons per layer. Similarly, the gating function in Visual Sudoku incorporates the same convolutional encoder, whose output is connected to a three-layer fully connected network with 128 neurons per layer. The learning rates for the actor, critic, and gating networks are set to $3\times10^{-4}$, $3\times10^{-4}$, and $2\times10^{-4}$, respectively. The gating function is trained using the Adam optimizer with a batch size of 128, updated every 1,000 time steps with 1,000 gradient updates per interval. Other hyperparameters follow the standard PPO settings\footnote{The code of NSAM is open-sourced at: https://github.com/shan0126/NSRL}.

\subsection{Baselines}

We compare ASG with representative baselines from four categories. (1) Classical RL: Rainbow \cite{Rainbow} and PPO \cite{schulman2017ppo}, two standard DRL methods, where Rainbow integrates several DQN enhancements and PPO is known for robustness and sample efficiency. (2) Neuro-symbolic RL: KCAC \cite{KCAC}, which incorporates domain constraint knowledge to reduce invalid actions and represents the state-of-the-art in action-constrained DRL. (3) Action masking: PLPG \cite{PLPG}, a state-of-the-art method that learns soft action masks for automatic action filtering. (4) Cost-based safe DRL: PPO-Lagrangian \cite{PPOlarg} and RC-PPO \cite{RCPPO}, which jointly optimize rewards and costs under the CMDP framework. For these methods, we construct the cost function using $(s,a,s',y)\in\Gamma_{\phi}$, where $y=0$ increases the cost by 1 \cite{wachi2023safe}.

\subsection{Learning efficiency and final performance}

To answer Q1, we compare our method with all baselines on 16 tasks across four domains. The learning curves are shown in Figure \ref{fig:compare}. The error bounds (i.e., shadow shapes) indicate the upper and lower bounds of the performance with 5 runs using different random seeds. During the learning process of NSAM, we apply the normalization process defined in Eq.~(\ref{ASG:equ_actionmask}), which plays a critical role in excluding unsafe or unexplorable actions from the RL policy. On the Sudoku tasks, as the size increases and the action-state space grows, NSAM exhibits a slight decrease in sample efficiency but consistently outperforms all baselines. A similar trend is observed in the N-Queens domain as the number of queens increases. In the Graph Coloring tasks, NSAM is able to converge to the optimal policy regardless of the graph topology among Graph 1$\sim$4. For the Visual Sudoku tasks, NSAM shows small fluctuations in performance after convergence due to the occurrence of previously unseen images. These fluctuations remain minor, indicating that NSAM’s symbolic grounding and learned policy generalize well to unseen digit images. The final converged reward values are reported in Table~1. Overall, NSAM achieves stable and competitive performance, consistently matching or outperforming the baselines.


\subsection{Less violation}


To answer Q2, we record constraint violations during training for each task. An episode is considered to violate constraints if any transition $(s,a,s',y)$ is labeled with $y=0$, i.e., the action $a$ is not explorable in that state. The violation rate is computed as the ratio of the episodes violating constraints to the total number of episodes. The final violation rates are summarized in Table~1. MSAM achieves significantly lower violation rates than all other methods.

\begin{figure}[t]
	\centering
	\subfigure[Sudoku 2 $\times$ 2]{\includegraphics[width=3.8cm]{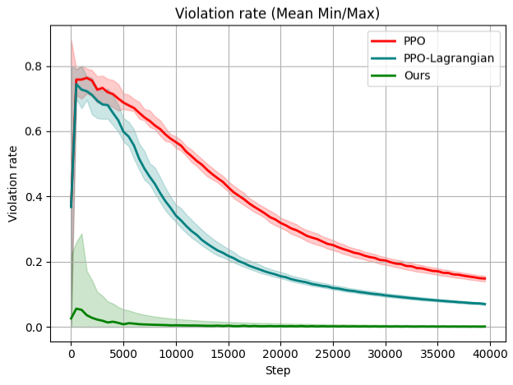}}
	\subfigure[Sudoku 3 $\times$ 3]{\includegraphics[width=3.8cm]{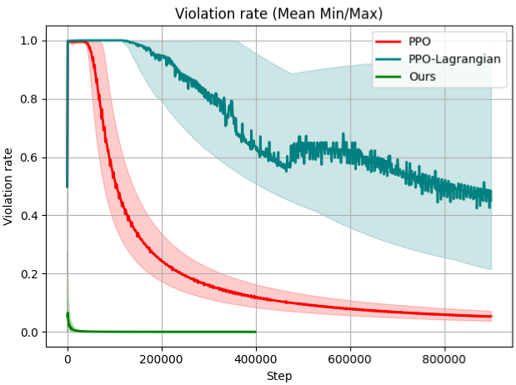}}
	\vspace{-0.4cm}
	\caption{Violation rate during training}
	\label{fig:viol}
\end{figure}

\begin{figure}[t]
	\centering
	\subfigure[Sudoku 3 $\times$ 3]{\includegraphics[width=3.8cm]{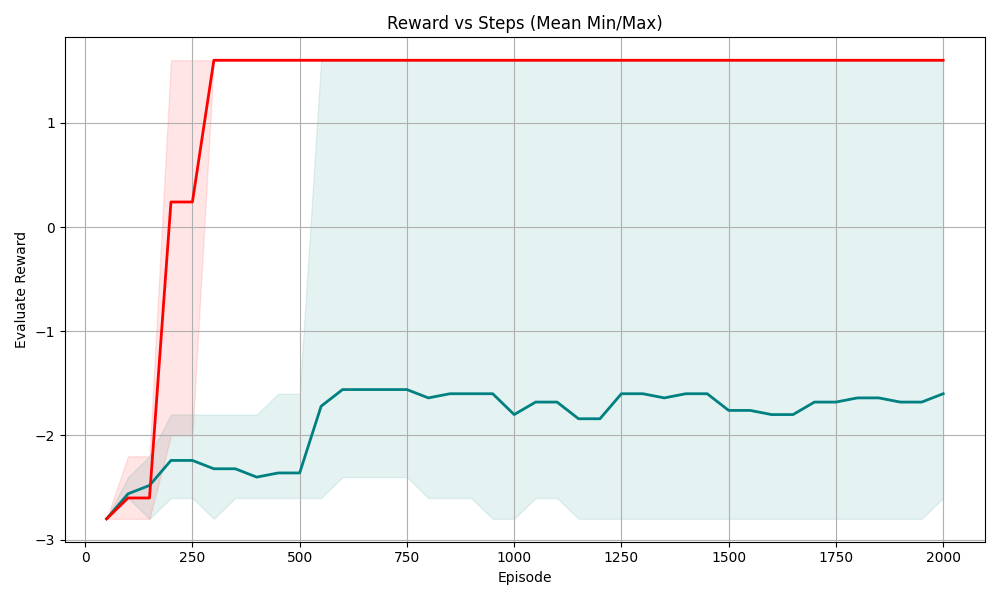}}
	\subfigure[Sudoku 4 $\times$ 4]{\includegraphics[width=3.8cm]{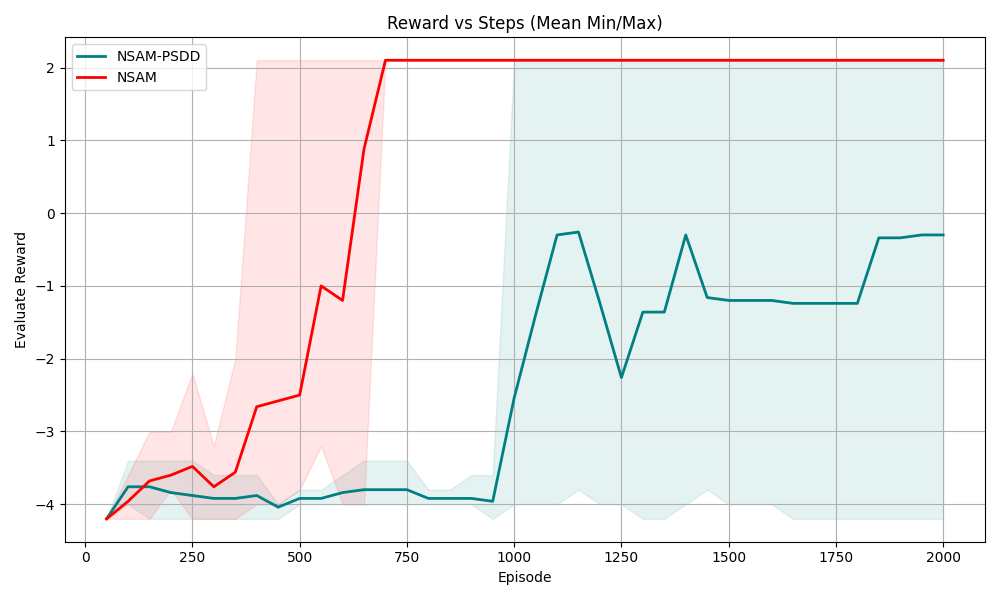}}
	\vspace{-0.4cm}
	\caption{Ablation study on PSDD in NSAM}
	\label{fig:ab_NSAM}
\end{figure}

We further compare the change of violation rates during training for NSAM, PPO, and PPO-Lagrangian, as shown in Figure \ref{fig:viol}. In the early stages of training, NSAM exhibits a slightly higher violation rate because the PSDD parameters is not trained, which may cause inaccurate evaluation of action preconditions $\varphi$. However, as training progresses, the violation rate rapidly decreases to near zero. During the whole training process, the violation rate of NSAM is consistently lower than that of PPO and PPO-Lagrangian.

\subsection{Ablation study}


To answer Q3, we conduct an ablation study on the symbolic grounding module by replacing the PSDD with a standard three-layer fully connected neural network (128 neurons per layer). The experiment results are shown in Figure \ref{fig:ab_NSAM}. Unlike PSDDs, neural networks struggle to efficiently exploit symbolic knowledge and cannot guarantee logical consistency in their predictions. As a result, the policy trained with this ablated grounding module exhibits highly unstable performance, confirming the importance of PSDD for reliable symbolic grounding in NSAM.

\subsection{Exploiting knowledge structure}

\begin{figure}[t]
	\centering
	\includegraphics[width=0.7\linewidth]{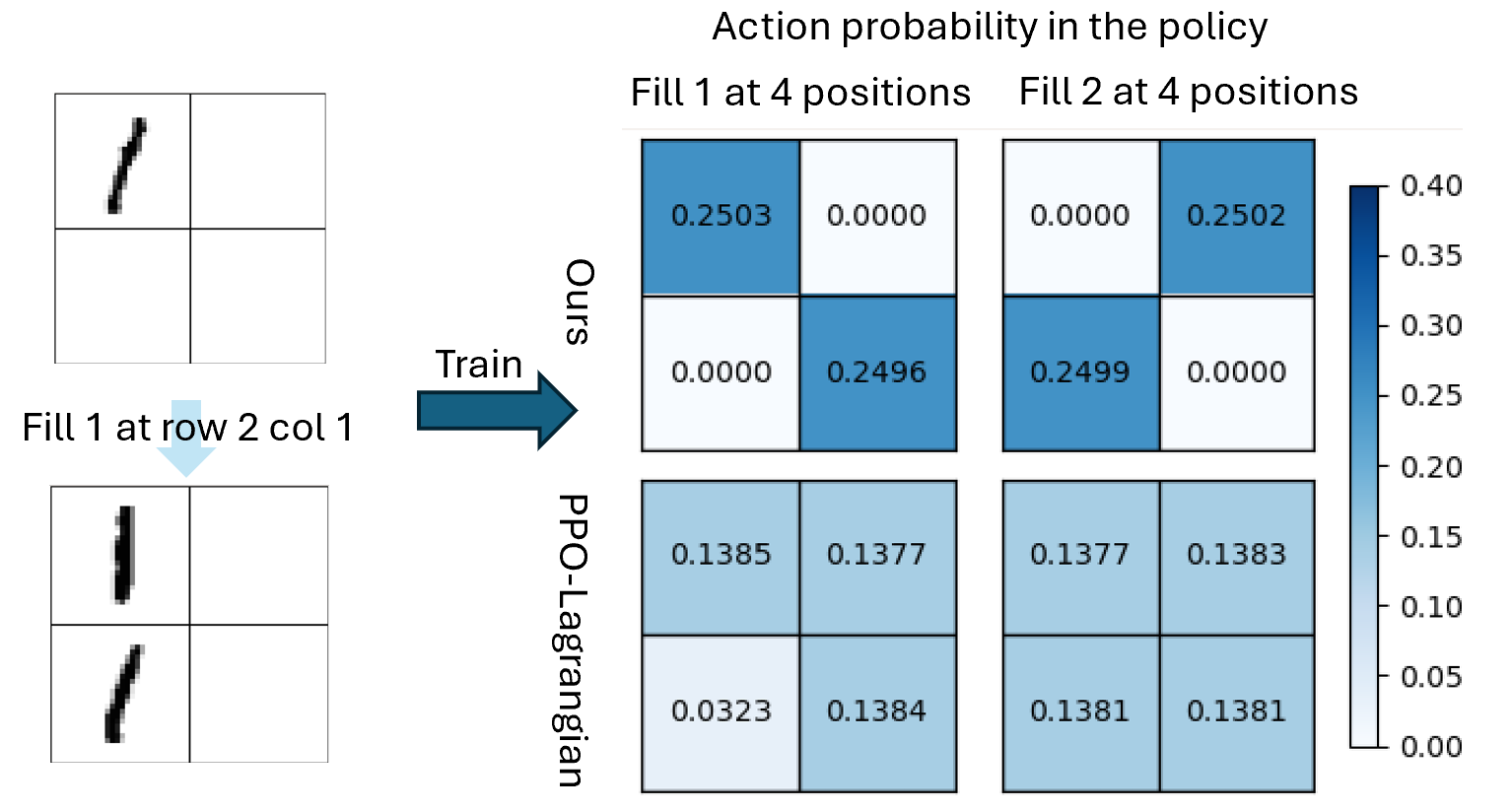}
	\vspace{-0.3cm}
	\caption{Policy training result from a single transition.}
	\label{fig:1data}
\end{figure}

To answer Q4, we design a special experiment where NSAM and PPO-Lagrangian are trained using only a single transition, as illustrated on the left side of Fig.~\ref{fig:1data}. The right side of the figure shows the heatmaps of action probabilities after training. With the cost function defined via $\Gamma_{\phi}$, PPO-Lagrangian can only leverage this single negative transition to reduce the probability of the specific action in this transition. As a result, in the heatmap of action probabilities after training, only the probability of this specific action decreases. In contrast, our method can exploit the structural knowledge of action preconditions to infer from the explorability of one action that a set of related actions are also infeasible. Consequently, in the heatmap, our method simultaneously reduces the probabilities of four actions. This demonstrates that our approach can utilize knowledge to generalize policy from very limited experience, which can significantly improve sample efficiency of DRL.

\section{Conclusions and future work}

In this paper, we proposed NSAM, a novel framework that integrates symbolic reasoning with deep reinforcement learning through PSDDs. NSAM addresses the key challenges of learning symbolic grounding from high-dimensional states with minimal supervision, ensuring logical consistency, and enabling end-to-end differentiable training. By leveraging action precondition knowledge, NSAM learns effective action masks that substantially reduce constraint violations while improving the sample efficiency of policy optimization. Our empirical evaluation across four domains demonstrates that NSAM consistently outperforms baselines in terms of sample efficiency and violation rate. In this work, the symbolic knowledge in propositional form. A promising research direction for future work is to investigate richer forms of symbolic knowledge as action preconditions, such as temporal logics or to design learning framework when constraints are unknown or incorrect. Extending NSAM to broader real-world domains is also an important future direction. 



\begin{acks}
We sincerely thank the anonymous reviewers. This work is partly
funded by the China Scholarship Council (CSC).
\end{acks}



\bibliographystyle{ACM-Reference-Format} 
\balance
\bibliography{sample}


\end{document}